\documentclass[review]{elsarticle}%单
\pdfoutput=1
\biboptions{numbers,sort&compress}
\usepackage[numbers,sort&compress]{natbib}
\usepackage{lineno}

\usepackage{tabularx,booktabs}
\usepackage{geometry}
\usepackage{multirow}
\usepackage{graphicx} %提供网格缩放
\usepackage{float}%控制图片位置
\usepackage{setspace}% 调整行间距
\usepackage{multicol}%分栏
\usepackage{geometry}%设置页边距
\usepackage{fontawesome}%特殊符号,以\fa开头
\usepackage{multirow}%合并行
\usepackage{makecell}%合并行
\usepackage{array}%设置表格行
\usepackage{bm} %公式加粗
\usepackage{enumitem}
\usepackage{amsfonts}
\usepackage{amsmath}%公式对齐工具
\usepackage{subfigure}%子图
\usepackage{mathtools} 
\journal{Journal of \LaTeX\ Templates}
\usepackage{caption}
\usepackage[ruled]{algorithm2e}
\usepackage{hyperref}
\hypersetup{hypertex=true,
	colorlinks=true,
	linkcolor=blue,
	anchorcolor=blue,
	citecolor=blue}
\begin{document}
\captionsetup[figure]{labelfont={bf},labelformat={default},labelsep=period,name={Fig.}}
% \captionsetup[table]{labelfont={bf},labelsep=newline,singlelinecheck=false}
%\titleformat{\subsection}[block]{\LARGE\bfseries}{Sec.\arabic{section}}{1em}
\begin{frontmatter}

\title{A physics-driven sensor placement optimization methodology for temperature field reconstruction}
% \title{Physic field reconstruction using physics-driven sensor placement methodology}


\author[mymainaddress,mymainaddress4]{Xu Liu}
\ead{liuxu18054448691@126.com}
\author[mymainaddress,mymainaddress4]{Wen Yao\corref{mycorrespondingauthor}}
\ead{wendy0782@126.com}
\author[mymainaddress,mymainaddress4]{Wei Peng\corref{mycorrespondingauthor}}
\ead{weipeng0098@126.com}
\author[mymainaddress2,mymainaddress3]{Zhuojia Fu}
\ead{zhuojiafu@gmail.com}
\cortext[mycorrespondingauthor]{Corresponding author}
\author[mymainaddress,mymainaddress4,mymainaddress5]{Zixue Xiang}
\ead{xiangzixuebit@163.com}
\author[mymainaddress,mymainaddress4]{Xiaoqian Chen}
\ead{chenxiaoqian@nudt.edu.cn}
% \author[mymainaddress]{Xiaoqian Chen}

% \author[mymainaddress]{Zixue Xiang}

\address[mymainaddress]{Defense Innovation Institute, Academy of Military Science, Beijing 100071, China}
\address[mymainaddress4]{Intelligent Game and Decision Laboratory, Academy of Military Science, Beijing 100071, China}
\address[mymainaddress5]{College of Aerospace Science and Engineering, National University of Defense Technology, Changsha 410073, China}
\address[mymainaddress2]{Laboratory of Ministry of Education for Coastal Disaster and Protection, Hohai
University, Nanjing 210098, China}
\address[mymainaddress3]{Center for Numerical Simulation Software in Engineering and Sciences, College of
Mechanics and Materials, Hohai University, Nanjing 211100, China}
\begin{abstract}
% Sensor placement optimization is a central challenge in physical field reconstruction. 
% Perception of the global field from sparse sensors is an essential technology,  to support the monitoring, analysis, and design of physical systems

 Perceiving the global field from sparse sensors has been a grand challenge in the monitoring, analysis, and design of physical systems. In this context, sensor placement optimization is a crucial issue. Most existing works require large and sufficient data to construct data-based criteria, which are intractable in data-free scenarios without numerical and experimental data. To this end, we propose a novel physics-driven sensor placement optimization (PSPO) method for temperature field reconstruction using a physics-based criterion to optimize sensor locations. In our methodological framework, we firstly derive the theoretical upper and lower bounds of the reconstruction error under noise scenarios by analyzing the optimal solution, proving that error bounds correlate with the condition number determined by sensor locations. Furthermore, the condition number, as the physics-based criterion, is used to optimize sensor locations by the genetic algorithm. Finally, the best sensors are validated by reconstruction models, including non-invasive end-to-end models, non-invasive reduced-order models, and physics-informed models. Experimental results, both on a numerical and an application case, demonstrate that the PSPO method significantly outperforms random and uniform selection methods, improving the reconstruction accuracy by nearly an order of magnitude. Moreover, the PSPO method can achieve comparable reconstruction accuracy to the existing data-driven placement optimization methods.

% To this end, we propose a novel physics-driven sensor placement optimization (PSPO) method for the physical field reconstruction using a physics-based criterion the as objective function. The mathematical model of the physical field is first discretized. Then, the reconstruction task with measurements is transformed into an unconstrained optimization problem by the Lagrange multiplier method. Furthermore, under noise scenarios, the theoretical upper and lower bounds of the reconstruction error are derived. It is theoretically proven that upper and lower bounds are positively correlated with the condition number of the coefficient matrix determined by the sensor locations. Thirdly, the condition number is used as the criterion to optimize sensor locations through the genetic algorithm. Finally, the optimized sensors are utilized in reconstruction models including non-invasive end-to-end models, non-invasive reduced-order models and physics-informed models, to reconstruct the physical field. 

%请用英语修改上面表述和语法更符合科技论文写作，并且检查内容是否合适
\end{abstract}


\end{frontmatter}

% \linenumbers

\section{Introduction}

% Reconstruction physical fields from sparse and noisy measurements have received attention in performance prediction \cite{zhang2019real,zhao2019power}, state  control \cite{yang2017multi, de2022sensing}, and health monitoring \cite{elodie2020sensors, zio2022prognostics} of physical systems governed by partial differential equations (PDEs). 
%\linenumbers
In performance prediction, state control, and health monitoring of physical systems, reconstructing the global field (e.g., temperature field) from sparse and noisy measurements is a long-stand challenge \cite{zhang2024simultaneous, de2022sensing, lu2023three, zio2022prognostics}. Sensor placement optimization plays a pivotal role in enhancing the accuracy of reconstruction models. Generally, two potential sources of information are available: 1) the mathematical model of the physical system. However, this model is often incomplete due to the parameter uncertainty or simplified due to the overcomplexity of the system; 2) experimental measurements obtained from sensors, which are often noise and limited to sparse locations within the domain \cite{introini2023stabilization}. In this content, there are two main topics of particular interest, reconstruction models and sensor placement optimization.

% State estimation from sparse and noisy measurements is of fundamental importance in terms of performance prediction, system control, prognostics, and health management, especially for the physics system governed by partial differential equations (PDEs). The physical field reconstruction of the physics system is a critical focus in many physics scientific and engineering applications. Generally, two possible sources of information can be available: 1) the mathematical model of the physics system. However, it is often incomplete due to the uncertainty of the parameters or simplified due to the overcomplexity of the system; 2) experimental measurements coming from sensors, which are prone to measurement noise and restricted to sparse locations within the domain. There are two main aspects of particular interest in the 

% State estimation from sparse and noisy measurements is a critical focus on performance prediction, system control, prognostics, and health management, especially for the physics system governed by partial differential equations (PDEs). Generally, two possible sources of information can be available: 1) the mathematical model of the physics system. However, it is often incomplete due to the uncertainty of the parameters or simplified due to the overcomplexity of the system; 2) experimental measurements coming from sensors, which are prone to measurement noise and restricted to sparse locations within the domain.
% 请用英语修改上面表述和语法更符合科技论文写作，并且检查内容是否合适

The first topic pertains to the models used for reconstructing the global field based on experimental measurements. Reconstructing the global field is a challenging task due to its high dimensionality, often involving tens of thousands or even millions of dimensions. Recently, researchers have explored neural network-based modeling techniques to address this high-dimensional mapping problem and overcome the challenges of the “curse of dimensionality"  \cite{bach2017breaking, abiodun2018state, cioica2022deep, fukami2021global}. These modeling techniques can mainly be categorized into three types. The first one is non-invasive end-to-end models, which utilize the neural network (NN) to learn the mapping relationship from measurements to global physical fields. The second one is non-invasive reduced-order models, which reduce the global fields to low-dimensional spaces and use measurements to determine low-dimensional spatial coefficients. The third one is physics-informed models, incorporating physical laws into the loss function to constrain the feasible solutions in the form of soft constraints

Extensive research has recently been conducted on non-invasive end-to-end models using popular NNs, such as multilayer perceptron (MLP) \cite{williams2022data, guastoni2020use}, convolutional neural network (CNN) \cite{peng2022deep, liu2021supervised, wan2022real}, graph neural network (GNN) \cite{duthe2023graph}, and deep operator network (DeepOnet) \cite{lu2021learning, deng2022approximation, goswami2022deep}. Despite these end-to-end models achieve high approximation accuracy, they still lack physical interpretation. Thus, researchers have turned to combining NNs with non-invasive reduced-order models. One popular approach is MLP-based reduced order models, which learn the mapping from measurements to Proper Orthogonal Decomposition (POD) coefficients \cite{erichson2020shallow, giannopoulos2020data, jiang2022online}. 
% This approach reduce the high-dimensional physical fields to the low-dimensional space represented by POD modes and coefficients. 
% Compared with POD-based models that ignore the temporal dynamics, Dynamic Mode Decomposition (DMD) \cite{guo2019data} explicitly extracts temporal dynamics and identifies coherent structures over time, providing a more accurate representation of dynamics systems. 
% Therefore, some studies \cite{nakamura2021convolutional, zhang2022unsteady} have combined DMD with long short-term memory (LSTM), where LSTM models dynamic evolution of measurements and DMD extracts coherent spatiotemporal structures.
% Recent advances in reduced-order models have achieved high accuracy. 
However, reduced-order models require significant training data to establish POD modes and do not explicitly incorporate physics constraints, such as partial differential equations and boundary conditions \cite{koo2020model, jo2019effective}. To address these limitations, researchers have resorted to physics-informed models, particularly physics-informed neural networks (PINN) \cite{arzani2021uncovering, liu2022novel, ouyang2023reconstruction}. PINN can reduce its dependence on the amount of measurements by incorporating physics constraints. These NN-based modeling techniques show superiority in addressing the challenges associated with reconstruction problems, offering efficient and accurate solutions. Moreover, optimal sensor placement is a critical factor in improving the efficiency and accuracy of reconstruction models.

The second topic concerns sensor placement optimization, which defines and optimizes the performance criterion of sensor locations. Typically, a typical sensor placement optimization method consists of the objective function (performance criterion) and the optimization technique. The criteria are utilized to evaluate sensor locations and are combined with optimization algorithms to solve the sensor placement optimization problem \cite{cheng2010thermal, domingo2016sensor, li2021efficient}. There are mainly three types of optimization algorithms, including convex relaxation methods, heuristic methods, and greedy methods. Since the sensor placement optimization problem is often non-convex, convex relaxation methods are used to approximate the original non-convex optimization problem through a convex relaxation, but the computational cost remains relatively high for large-scale problems \cite{shchetinin2019efficient, blachowski2020convex}. Recently, the method based on non-convex optimization using the proximal splitting algorithm has been developed for handling the large-scale problem \cite{nagata2021data}. Heuristic methods utilize predefined rules, guidelines, or experiences from domain knowledge to guide sensor placement \cite{kong2022optimal}. Genetic algorithms (GA), particle swarm optimization, and ant colony optimization are common heuristic methods. Furthermore, greedy methods, a special class of heuristics algorithms, sequentially select sensor locations based on immediate local optimization without considering global optimality. Greedy selection or greedy elimination algorithms have been extensively studied \cite{nagata2022data, nagata2023randomized}.
Moreover, the column-pivoted QR decomposition algorithm has recently drawn increasing attention \cite{zhou2020temperature, manohar2018data, clark2020multi}. Generally, heuristic and greedy methods are computationally efficient and easy to implement, but they may lead to sub-optimal solutions and lack global optimization guarantees. More importantly, criteria are the objective of optimization algorithms, guiding the direction of sensor placement. Some studies consider error-based criteria to directly minimize the difference between predictions and exact values \cite{clark2018greedy}. However, sensor placement with error-based criteria is a combinatorially hard problem and is usually intractable. Therefore, most researchers have focused on exploring information-theoretic and reduced model-based criteria \cite{clark2020multi, li2021efficient}.

Information-theoretic criteria are designed to maximize the information or minimize the uncertainty associated with the estimated quantities of interest. These criteria take into account various factors, including the spatial distribution of sensors, and the relationship between measurements and the target variables. One commonly used information-theoretic criterion is mutual information (MI), which quantifies the information that sensors provide about the target variables and identifies the most informative sensor locations \cite{bhattacharyya2020exploiting}. Another criterion used in sensor placement optimization is entropy to evaluate the spatial distribution of sensors and their effectiveness in reducing uncertainty or capturing important system features \cite{christodoulou2013entropy}. 
% By maximizing the entropy criterion, sensor placement optimization aims to capture the most informative and diverse aspects of the system. 
Information-theoretic criteria provide quantitative measures but rely on specific assumptions about underlying probability distributions and lack interpretability regarding the underlying physical laws.

Model-based criteria provide an effective approach to evaluating sensor locations. textcolor{red}{In this framework, the global physical fields are generally characterized by modes and coefficients.} Sensor placement optimization seeks optimal modes to accurately estimate coefficients, resulting in improving reconstruction performance. Various criteria have been explored to quantify the dual measurements matrix determined by modes. One commonly used criterion is A-Optimality to efficiently estimate coefficients by minimizing the trace of the measurements matrix. Maximizing Fisher information of the measurements matrix also leads to minimizing the trace of the measurements matrix \cite{yamada2021fast, jana2019fisher}. Other used criteria include D-optimality and the E-optimality criterion, which maximizing the determinate of the measurements matrix and minimizing the maximum eigenvalue of the measurements matrix, respectively. Moreover, minimizing the condition number of the measurements matrix is essential for reducing noise amplification \cite{manohar2018data}, making it a critical factor in sensor placement optimization \cite{yildirim2009efficient, sun2018wind}.

Despite the advances in data-driven criteria, they often require large datasets and lack interpretability in terms of physical laws, which are intractable in data-free scenarios without numerical and experimental data. For instance, in the design phase, the sensor locations have to be determined before acquiring experimental measurements. To this end, we propose a physic-driven sensor placement optimization (PSPO) method for temperature field reconstruction, where a physics-based criterion is derived, and GA is used to optimize the criterion. Concretely, we first discretize the mathematical model by the radial basis function finite difference technique. Then, the reconstruction task with measurements is modeled as an unconstrained optimization problem by the Lagrange multiplier method. Moreover, under the noise scenarios, the theoretical upper and lower bounds of the reconstruction error are derived by analyzing the optimal solution. Theoretical analysis demonstrates that upper and lower bounds correlate with the condition number of the coefficient matrix, which is uniquely determined by sensor locations. Secondly, minimizing condition number as the criterion is used to optimize the sensor locations by GA. Finally, the best sensors are validated by reconstruction models, including non-invasive end-to-end models, non-invasive reduced-order models, and physics-informed models. In summary, our contributions are as follows:
\begin{itemize}
    \item A physics-based criterion is proposed. The upper and lower bounds of reconstruction errors are analyzed theoretically, proving which are related to the condition number determined by sensor locations. Thus, the condition number is used physics-based criterion to evaluate sensor locations.
    \item A physics-driven sensor placement optimization (PSPO) method is proposed for temperature field reconstruction based on the physics-based criterion and GA, validated by non-invasive end-to-end, non-invasive reduced-order, and physics-informed models.
    \item Experimental results validate the feasibility and effectiveness of the PSPO method. Compared with random and uniform selection methods, the PSPO method provides superior sensor locations to improve reconstruction performances of models. Besides, the reconstruction performances by the PSPO method are comparable to those by the existing data-driven placement optimization methods.
\end{itemize}

The remainder of this paper is organized as follows: Section \ref{sec problem} presents the problem formulation. Section \ref{Methodology} details the PSPO method and briefly reviews the validation of the proposed method. Numerical analysis is discussed in Section \ref{sec result}, highlighting the results of the PSPO method in both a numerical case and an application case. Section \ref{sec conclusion} concludes the paper.

\section{Problem Setup}
\label{sec problem}
In engineering applications, reconstructing the temperature field is critical for ensuring safety, enabling effective control, and facilitating supervision of physical systems. In this regard, two possible information sources are generally available: the mathematical model and experimental measurements from sensors. However, mathematical models may be incomplete or have limitations due to uncertainties about the parameters, and measurements are often affected by random and correlated noise. 

Given the partial mathematical model of the physical system, the temperature field reconstruction (TFR) task can be formulated as an $L_{2}$ minimization problem constrained by the following equation:

\begin{equation}
\label{eq_r_task}
\begin{aligned}
& \underset{\hat{\boldsymbol{u}}}{\min} \left\|\boldsymbol{u}\left(\boldsymbol{x}_{loc}\right)-\hat{\boldsymbol{u}}\left(\boldsymbol{x}_{loc}\right)\right\|^2, \\
& \operatorname{s.t.} \mathcal{L}\hat{\boldsymbol{u}}(\boldsymbol{x})= f(\boldsymbol{x};\boldsymbol{\lambda}), in  \,  \Omega, \\
& \qquad \mathcal{B}(\hat{\boldsymbol{u}}(\boldsymbol{x}))=h, on \, \partial \Omega,
\end{aligned}
\end{equation}
where $\hat{\boldsymbol{u}}$ denotes the reconstructed temperature field,  $\boldsymbol{u}\left(\boldsymbol{x}_{loc}\right)$ denotes measurements, $\boldsymbol{x}_{loc} \in \Omega$ denotes the sensor locations. The constraint has to satisfy physical laws, including the partial differential equation (PDE) and the boundary condition (BC). $\mathcal{L}$ denotes a Laplace operator. $f(\boldsymbol{x},\boldsymbol{\lambda})$ and $h$ denote the source term with parameter $\boldsymbol{\lambda}$ and boundary value, respectively. $f(\boldsymbol{x},\boldsymbol{\lambda})$ and $h$ are defined as unknown problem parameters in this work. $\mathcal{B}$, $\Omega$, and $\partial \Omega$ denote the boundary condition, the domain, and the boundary area, respectively.

% It is noteworthy that due to uncertainties about problem parameters, solving the physical system using numerical methods such as finite difference and finite element methods can be challenging. Therefore, various reconstructed models by measures have been developed to obtain reconstructed physics field.

% where $\hat{u}$ is the reconstructed physical field, $u\left(\boldsymbol{x}_{loc}\right)$ is the measurements, $\boldsymbol{x}_{loc}$ is the sensor location, and $\mathcal{N}$ is a nonlinear operator with parameter $\lambda$, which is unknown in this work. $\Delta u$ represents the gradient in space, $\Delta u$ is the Laplace item, which is equal to $\nabla_{2} u$. $\mathcal{B}$ represents the boundary condition, and $\partial \Omega$ denotes the boundary area of $\Omega$. It is noteworthy that due to uncertainties about parameter values, solving physical systems directly using numerical methods such as finite difference and finite element methods can be challenging.

In the TFR task, sensor placement optimization plays a critical role since it seeks sensor locations to capture or maximize relevant information, thereby alleviating the impact of noise and enhancing the accuracy of the model. Let the index set of candidate locations and the selected sensor locations be denoted by $\mathcal{V}=\{\mathcal{S}_{1}, \cdots, \mathcal{S}_{n}\}$ and $\boldsymbol{x}_{loc}=\left\{\boldsymbol{x}_1, \cdots, \boldsymbol{x}_k\right\}$. The sensor placement problem for the reconstruction task can be mathematically formulated as
\begin{equation}
\underset{\boldsymbol{x}_{loc} \subseteq \mathcal{V},|\boldsymbol{x}_{loc}|=k}{\operatorname{argmin}} \mathrm{F}(\boldsymbol{x}_{loc}),
\end{equation}
where $\mathrm{F}(\boldsymbol{x}_{loc})$ represents the objective function (or criterion) used to evaluate reconstruction performances. While using reconstruction error as the criterion is possible, this approach results in a combinatorial optimization problem, which is NP-hard and computationally intractable, even for moderately large problems. Therefore, establishing the criterion of sensor locations to evaluate reconstruction performances remains a long-term challenge.
% The sensor placement task generally is modeled as
% \begin{equation}
% \label{optima_problem}
% \left\{\begin{array}{l}
% \text { find } \boldsymbol{x}_{o b s} \\
% \min _{u} \left\|u\left(\boldsymbol{x}_{o b s}\right)-u_{o b s}\left(\boldsymbol{x}_{o b s}\right)\right\|^2 \\
% \text { s.t. } \mathcal{N}[u, \cdots, \nabla u, \Delta u, \nabla u \cdot u, \cdots ; \lambda]=\mathbf{0} \\
% \mathcal{B}(u, \nabla u, \Delta u, \cdots ; \boldsymbol{x} \in \partial \Omega)=\mathbf{0}
% \end{array}\right.
% \end{equation}

In this work, we consider the noise scenario, where measurements are assumed to be independently Gaussian distributed, centered around the hidden exact value $\boldsymbol{u}^{true}$. Mathematically, this can be expressed as
\begin{equation}
\boldsymbol{u}(\boldsymbol{x}_{loc})=\boldsymbol{u}^{true}(\boldsymbol{x}_{loc})+\epsilon \boldsymbol{u}^{true}(\boldsymbol{x}_{loc}),
\end{equation}
where $\epsilon$ represents independent Gaussian noise with zero mean. The fidelity of each sensor is known, and the standard deviation of $\epsilon$ is denoted by $\sigma$. Gaussian noise is commonly used in modeling because it has a well-defined statistical distribution, is mathematically tractable, and often reflects real-world noise characteristics due to the Central Limit Theorem. Therefore, Gaussian noise is the primary focus of our study. Although we only considered Gaussian noise, the current study serves as an important foundation for future exploration of related noise \cite{liu2016sensor}.

The core focus of our research lies in the design of sensor placement under conditions of data scarcity, specifically in data-free scenarios where there is an absence of ample simulation or experimental data to draw upon. This assumption stems from our engineering practices and represents a real-world problem that demands a solution. For example, in spacecraft engineering design, the design time allocated for sensor placement is extremely limited. According to our engineering experience, urgent tasks typically allow only 1-2 days for placement design. Despite having access to a complete simulation model, generating the requisite global temperature field data is time-consuming, which significantly exceeds the design time of placement. Therefore, these practical task requirements compel us to rapidly provide sensor placement strategies under data-free scenarios.

\section{Methodology}
\label{Methodology}

\subsection{Overview}

Aiming at data-free scenarios, we propose a physic-driven sensor placement optimization method for temperature field reconstruction. This method utilizes a physics-based criterion and GA, which is validated by non-invasive end-to-end models, non-invasive reduced-order models, and physics-informed models. In practical applications, based on the incomplete mathematical model, we propose the PSPO method to construct physics-based criteria and further utilize GA to iteratively update sensor placement during the placement phase. The detailed procedure is shown in Algorithm \ref{alg_PSPO}. Sensor placement and reconstruction learning can be regarded as two stages. In the reconstruction phase (validation of placement method), the global field dataset can be acquired through infrared cameras or numerical simulations. Based on determined placement, sensor observations are obtained from patch-type sensors deployed or through numerical simulations. Furthermore, NNs are trained to learn the mapping from sensor observations to the global fields. Fig. \ref{fig:PSPO} shows a schematic diagram of our approach. Details of the physic-driven sensor placement optimization method and the validation of the placement method will be presented as follows.

\subsection{The physics-driven sensor placement optimization method}
\begin{figure*}[!ht]
	\centering
\includegraphics[width=0.95\linewidth]{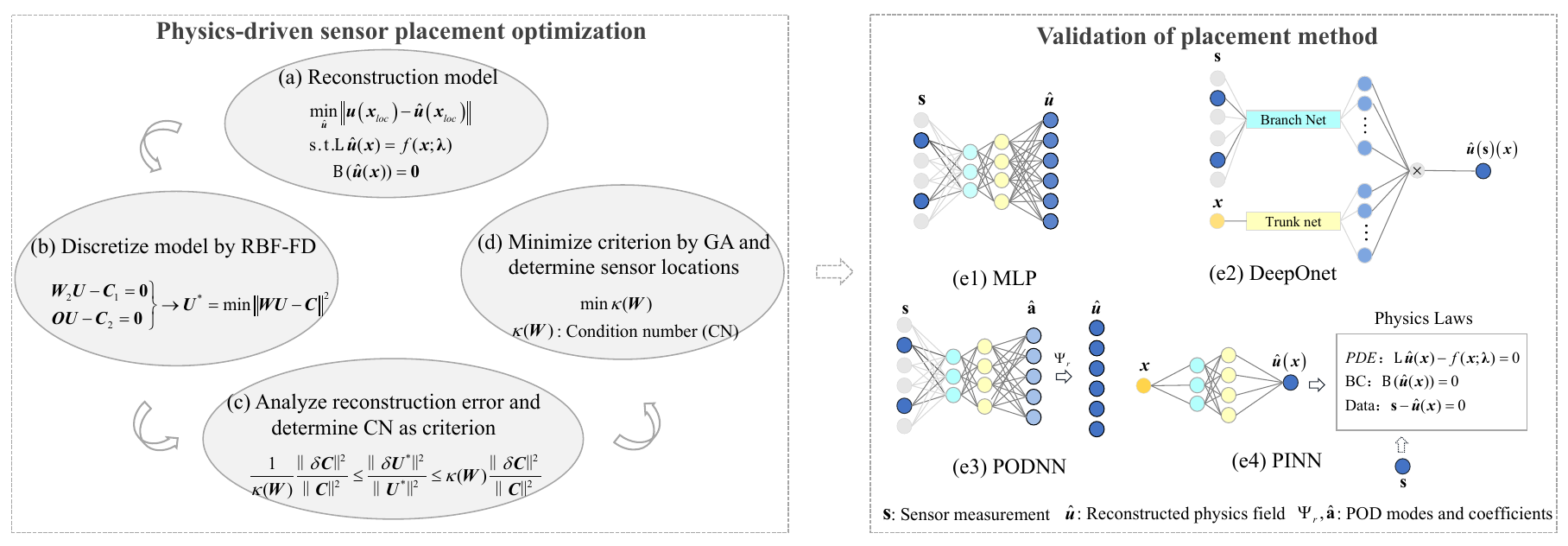}
	\caption{Conceptual flow of the PSPO method for temperature field reconstruction, where MLP and DeepOnet belong to non-invasive end-to-end models, PODNN belongs to the non-invasive reduced-order model, and PINN belongs to physics-informed model.}
	\label{fig:PSPO}
\end{figure*}

Most existing studies on the sensor placement problem have focused on established data-based criteria, such as model-based criteria and information-theoretic criteria. In data-free scenarios, these criteria may not be feasible, but the incomplete mathematical model is accessible, such as during the equipment design stage. To this end, we propose a physic-driven sensor placement optimization (PSPO) method, where a physics-based criterion is derived and GA is used to optimize the criterion. The conceptual flow of the PSPO method is illustrated in Fig. \ref{fig:PSPO}, and detailed steps are as follows. 

\begin{algorithm*}[!h]
	\caption{\textbf{The physics-driven sensor placement optimization method}}
	\label{alg_PSPO}
	\LinesNumbered
	\KwIn{The sensor number $k$, point stencils of RBF-FD $n$, the nearest neighbor point of RBF-FD $m$.
	}
	\KwOut{The best sensor locations} $\hat{O}_{optimal}$.
	// \textbf{Discretize the mathematical model by the RBF-FD method} \\
	Generate point stencils by sampling techniques.\\
	Select $m$ nearest neighbor points to form the corresponding local region for each $x_i$.\\
	Calculate the RBF-FD weight for each point $x_i$ by Eq.(\ref{eq_weight_u})-(\ref{eq_rbf_fdweight_point}).\\
    Discretize the constraint of the TFR task through the RBF-FD weight in Eq.(\ref{eq_constraint}).\\
    Represent the optimization objective through the location matrix in Eq.(\ref{eq_objective}).\\
    Transform the TFR task into an unconstrained optimization problem in Eq.(\ref{eq_uncons_task}).\\
	// \textbf{Determine the physics-informed criterion}\\
    Analyze error bounds of the optimal solution given a noisy scenario.\\
    Determine the criterion according to the Theorem 3.1.\\
    // \textbf{Optimize the criterion by GA}\\
    Minimize the physics-based criterion through GA based on an open-source library DEAP.\\
    Find the best locations $\hat{O}_{optimal}$.
\end{algorithm*}

\vspace{7em}

Firstly, we use the Radial Basis Function Finite Difference (RBF-FD) method to discretize the mathematical model, transforming the TFR task into an unconstrained optimization problem. RBF-FD is a mesh-free method that generalizes finite differences in such a way as to eliminate the dependency on Cartesian grids, thus enabling the solution of PDEs on point clouds \cite{shankar2015radial, shankar2017overlapped, su2019radial}. Assuming that there are $n$ point stencil $\left\{x_i\right\}_{i=1}^n$ within the domain, we select $m$ nearest neighbor points to form the corresponding local region $\Omega_i = \left\{x_i^j\right\}_{j=1}^m$ for each point $x_i$. The points can be generated using various sampling techniques, such as uniform sampling and random sampling. By using RBF-FD, we can approximate operator $\mathcal{L}$ acting on the function (i.e., to approximate $\mathcal{L}u$) at points $x_i$. This approximation can be obtained by a weighted linear combination of $u$ at $x_i$ and its $m$ nearest neighbor points, 
\begin{equation}
\label{eq_weight_u}
\left.\mathcal{L}u(x)\right|_{=x_i} \approx \sum_{j=1}^m w^{i}_j u\left(x_j\right),
\end{equation}
where $w^{i}_j$ denotes RBF-FD weights. The core concept of RBF-FD is to seek $w^{i}_j$. Polynomial functions $p\left(x_i\right)$ and radial basis functions $\phi\left(x_i, x_i^j\right)$ are used to estimate the solution $u\left(x_i\right)$, 
\begin{equation}
\label{eq_rbffd_u}
u\left(x_i\right) \approx \sum_{j=1}^m \eta_j \phi\left(x_i, x_i^j\right)+\sum_{j=1}^q \mu_j p_j\left(x_i\right),
\end{equation}
where $\eta_j$ and $\mu_j$ represents the corresponding coefficients. $q$ denotes the polynomial number with constraints $\sum_{j=1}^m \eta_j p_l\left(x_i\right)=0, l=1, \ldots, q$. It is simplified as
\begin{equation}
\label{eq_rbffd_u_simple}
\left[\begin{array}{l}
\boldsymbol{\eta} \\
\boldsymbol{\mu}
\end{array}\right]=\left[\begin{array}{cc}
\boldsymbol{A} & \boldsymbol{P} \\
\boldsymbol{P}^{\mathrm{T}} & \mathbf{0}
\end{array}\right]^{-1}\left[\begin{array}{l}
\boldsymbol{u} \\
\mathbf{0}
\end{array}\right],
\end{equation}
where $P_{i j}=p_j\left(x_i\right), i=1, \ldots, k, j=1, \ldots,q$, $A_{i j}=\phi(x_i,x_j), i, j=1, \ldots, m$, $\boldsymbol{u}=[u_1, u_2, \cdots, u_m]^T$.
By substituting Eq.(\ref{eq_rbffd_u}) into Eq.(\ref{eq_weight_u}), we can obtain  
\begin{equation}
\sum_{j=1}^k \eta_j \mathcal{L} \phi\left(x_i, x_i^j\right)+\sum_{j=1}^q \mu_j \mathcal{L} p_j\left(x_i\right) \approx \sum_{j=1}^m w^{i}_j u\left(x_j\right),
\end{equation}
which simplifies to
\begin{equation}
\label{eq_rbffd_bc}
\left[\begin{array}{ll}
\boldsymbol{b} & \boldsymbol{c}
\end{array}\right]\left[\begin{array}{l}
\boldsymbol{\eta} \\
\boldsymbol{\mu}
\end{array}\right]=\left[\begin{array}{ll}
\boldsymbol{w} & \boldsymbol{v}
\end{array}\right]\left[\begin{array}{l}
\boldsymbol{u} \\
\mathbf{0}
\end{array}\right],
\end{equation}
$b_{i j}=\mathcal{L}\phi(x_i,x_j), i, j=1, \ldots, k$, $c_{i j}=\mathcal{L}p_j\left(x_i\right), i=1, \ldots, k, j=1, \ldots,q$. $\boldsymbol{v}$ is Lagrange multipliers enforcing the condition $\boldsymbol{P}^{\mathrm{T}}\boldsymbol{w}^{k}=\boldsymbol{c}^{\mathrm{T}}$. By substituting Eq.(\ref{eq_rbffd_u_simple}) into Eq.(\ref{eq_rbffd_bc}), we can obtain the RBF-FD weight of $u(x_i)$ by the following equation,  
\begin{equation}
\label{eq_rbf_fdweight_point}
\left[\begin{array}{l}
\boldsymbol{w}^{\mathrm{T}} \\
\boldsymbol{v}^{\mathrm{T}}
\end{array}\right]=\left[\begin{array}{cc}
\boldsymbol{A} & \boldsymbol{P} \\
\boldsymbol{P}^{\mathrm{T}} & \mathbf{0}
\end{array}\right]^{-1}\left[\begin{array}{l}
\boldsymbol{b}^{\mathrm{T}} \\
\boldsymbol{c}^{\mathrm{T}}
\end{array}\right].
\end{equation}
Thus, we further obtain RBF-FD weights for every point $x_i, i=1, \ldots, n$.

Through RBF-FD, the constraint of the TFR task in Eq.(\ref{eq_r_task}) can be discretized as
\begin{equation}
\label{eq_constraint}
\boldsymbol{W}_1 \boldsymbol{U}_1 = f(\boldsymbol{x};\boldsymbol{\lambda}) + g(\boldsymbol{x}),
\end{equation}
where $\boldsymbol{W}_1$ denotes the RBF-FD weights and $\boldsymbol{U}_1=[u^1, u^2, \cdots, u^n]^T$ denotes the solution at points $\boldsymbol{x}=[x_1, x_2, \cdots, x_n]^T$, $g(\boldsymbol{x})$ is the vector constrained by the boundary condition, and source term $f(\boldsymbol{x};\boldsymbol{\lambda})$ can be written as $\sum_{j=1}^l \phi_j(\boldsymbol{x}) \lambda_j+\tilde{f}(\boldsymbol{x})$ by separating variables in this work. Eq.(\ref{eq_constraint}) is simplified as
\begin{equation}
\boldsymbol{W} \boldsymbol{U} - \boldsymbol{C}_1=\left(\boldsymbol{W}_1,-\boldsymbol{\phi} \right)\left(\begin{array}{l}
\boldsymbol{U}_1 \\
\boldsymbol{\lambda}
\end{array}\right)- \boldsymbol{C}_1= \mathbf{0}.
\end{equation}

Furthermore, the optimization objective can be expressed as
\begin{equation}
\label{eq_objective}
\min \left\|\boldsymbol{O} \boldsymbol{U}-\boldsymbol{C}_2\right\|^2=\min \left\|\left(\begin{array}{ll}
\boldsymbol{O}_1 & \boldsymbol{0}
\end{array}\right)\left(\begin{array}{l}
\boldsymbol{U}_1 \\
\boldsymbol{\lambda}
\end{array}\right)-\boldsymbol{C}_2\right\|^2,
\end{equation}
where $\boldsymbol{O}_1$ denotes the location matrix determined by $\boldsymbol{x}_{loc}$, and $\boldsymbol{C}$ is the measurement matrix. The matrix $\boldsymbol{O}_1$ is defined such that each row contains exactly one element equal to $1$, with all other elements in that row being $0$. For example, there are three sensors in $\boldsymbol{x}_1$, $\boldsymbol{x}_3$ and $\boldsymbol{x}_5$. $\boldsymbol{O}_1\boldsymbol{U}_1-\boldsymbol{C}$ can be expressed as
\begin{equation}
\left(\begin{array}{ccccccc}
1 & 0 & 0 & 0 & 0 & \cdots & 0 \\
0 & 0 & 1 & 0 & 0 & \cdots & 0 \\
0 & 0 & 0 & 0 & 1 & \cdots & 0
\end{array}\right)_{3 \times n} \quad \boldsymbol{
U}_{n \times 1}-\left(\begin{array}{c}
u_{o b s}^1 \\
u_{o b s}^3 \\
u_{o b s}^5
\end{array}\right)_{3 \times 1}.
\end{equation}

\vspace{7em}
Therefore, the TFR task with measurements in Eq.(\ref{eq_r_task}) is reformulated as an unconstrained optimization problem using the Lagrange multiplier method,
\begin{equation}
\label{eq_uncons_task}
\min \| \boldsymbol{W} \boldsymbol{U}-\boldsymbol{\boldsymbol{C}}\|^2,
\end{equation}
where $\boldsymbol{W}=(\gamma \boldsymbol{W}, \boldsymbol{O})^{T}$, $\boldsymbol{U} = (\boldsymbol{U}_1, \boldsymbol{\lambda})$, and $\boldsymbol{C} = (\gamma \boldsymbol{C}_1, \boldsymbol{C}_2)^T$, $\gamma$ denotes the Lagrange multiplier.

Secondly, we analyze error bounds of the optimal solution given a noisy scenario and determine the condition number as the physics-informed criterion. According to \textbf{Theorem 3.1}, we have proven that when the measurements $\boldsymbol{C}$ has a perturbation $\delta \boldsymbol{C}$, the TFR task has error upper and lower bounds, which is related to the condition number of $\boldsymbol{W}$. For
\begin{equation}
\boldsymbol{W} =\left(\begin{array}{c}
\gamma \boldsymbol{W}_1 \\
\boldsymbol{O}
\end{array}\right),
\end{equation}
if the mathematical model is well-defined, the matrix $\boldsymbol{W}_1$ can be uniquely determined. Moreover, the matrix $\boldsymbol{O}$ is determined by sensor locations. 
 
By strategically selecting these locations, it is possible to obtain a matrix $\boldsymbol{W}$ with a smaller condition number, which leads to a smaller upper bound and a larger lower bound on the error of the optimal solution. This results in a more accurate and precise solution. Therefore, we use the condition number of $\boldsymbol{W}$ as the 
physics-based criterion related to reconstruction performance.

Finally, we use GA to optimize sensor locations by minimizing the condition number (physics-based criterion). All implementations for GA are based on an open-source library DEAP \cite{kim2019software}. 

\noindent\textbf{Theorem 3.1} \textbf{(Relative error bound)}. Suppose that $\boldsymbol{W}$ is an $m \times n$ matrix with $m>n$ and has a full rank. Let $\boldsymbol{U}$ minimizes $\|\boldsymbol{W} \boldsymbol{U}- \boldsymbol{C}\|^2$ and $\boldsymbol{U}^{*}$ be the optimal solution. If there is a perturbation $\delta \boldsymbol{C}$ in $\boldsymbol{C}$, and $\boldsymbol{U}+\delta \boldsymbol{U}$ minimizes $\|\boldsymbol{W} (\boldsymbol{U}+\delta \boldsymbol{U})-(\boldsymbol{C}+\delta \boldsymbol{C})\|^2$ with $\boldsymbol{U}^{*}+\delta \boldsymbol{U}^{*}$ as the  corresponding solution, then,
\begin{equation}
\frac{1}{\kappa(\boldsymbol{W})} \frac{\|\delta \boldsymbol{C}\|}{\|\boldsymbol{C}\|} \leq \frac{\|\delta \boldsymbol{U}^{*}\|}{\|\boldsymbol{U}^{*}\|} \leq \kappa(\boldsymbol{W}) \frac{\|\delta \boldsymbol{C}\|}{\|\boldsymbol{C}\|},
\end{equation}
where $\kappa(\boldsymbol{W})$ presents the condition number of $\boldsymbol{W}$. $\kappa(\boldsymbol{W})$ is equivalent to $\|\boldsymbol{W}\|\left\|\left(\boldsymbol{W}^{H} \boldsymbol{W}\right)^{-1} \boldsymbol{W}^{H}\right\|$.

\noindent\textbf{Sketch of Proof of Theorem 3.1.} The derivation satisfies the following inequalities: 1) Matrix and vector norm must be compatible:$\|\boldsymbol{WU}\| \leq \|\boldsymbol{W}\| \|\boldsymbol{U}\|$. 2) We assume that $\| \boldsymbol{W}\|\left\|\left(\boldsymbol{W}^{H} \boldsymbol{W}\right)^{-1}\boldsymbol{W}^{H}\right\|\leq 1$. 

On the one hand, we assume $\boldsymbol{U}^{*}+\delta \boldsymbol{U}^{*}$ is the optimal solution under the noise scenario, 
\begin{equation}
\boldsymbol{W}(\boldsymbol{U}^{*}+\delta \boldsymbol{U}^{*})=\boldsymbol{C}+\delta \boldsymbol{C}.
\end{equation}
By the fact that $\boldsymbol{W}\boldsymbol{U}^{*}$ is equivalent to $\boldsymbol{C}$ under the noiseless scenario, we get that
\begin{equation}
\boldsymbol{W}\delta \boldsymbol{U}^{*}=\delta \boldsymbol{C}.
\end{equation}
Then, since $\left\|\delta \boldsymbol{U}^{*} \right\|=\left\|\boldsymbol{W}^{\dagger} \delta \boldsymbol{C}\right\|$, we can obtain rigorous absolute perturbation for $\boldsymbol{U}^{*}$, 
\begin{equation}
\label{eq_delta_U}
\|\delta \boldsymbol{U}^{*}\| \leq\|\boldsymbol{W}^{\dagger}\|\|\delta \boldsymbol{C}\|,
\end{equation}
where $\boldsymbol{W}^{\dagger}$ represents $\left(\boldsymbol{W}^{H} \boldsymbol{W}\right)^{-1} \boldsymbol{W}^{H}$. Furthermore, with 
 Eq.(\ref{eq_delta_U}) and $\|\boldsymbol{C}\|=\left\|\boldsymbol{W} \boldsymbol{U}^*\right\| \leq\|\boldsymbol{W}\|\|\boldsymbol{U}^{*}\|$, the upper bound of the first-order absolute error  for $\boldsymbol{U}^*$ is given by
\begin{equation}
\label{eq_upper}
\frac{\|\delta \boldsymbol{U}^{*}\|}{\|\boldsymbol{U}^{*}\|} \leq \kappa(\boldsymbol{W}) \frac{\|\delta \boldsymbol{C}\|}{\|\boldsymbol{C}\|},
\end{equation}
where $\kappa(\boldsymbol{W})$ is equivalent to $\|\boldsymbol{W}\|\left\|\left(\boldsymbol{W}^{H} \boldsymbol{W}\right)^{-1} \boldsymbol{W}^{H}\right\|$.

On the other hand, again by $\|\delta \boldsymbol{C}\|=\left\|\boldsymbol{W} \delta\boldsymbol{U}^*\right\| \leq\|\boldsymbol{W}\|\|\delta\boldsymbol{U}^*\|$, we have
\begin{equation}
\label{eq_delta_u_upper}
\frac{\left\|\delta \boldsymbol{U}^*\right\|}{\left\|\boldsymbol{U}^*\right\|} \geq \frac{\|\delta \boldsymbol{C}\|}{\| \boldsymbol{W}\|\left\|\boldsymbol{U}^*\right\|}.
\end{equation}
Combining Eq.(\ref{eq_delta_u_upper}) and $\left\|\boldsymbol{U}^*\right\|\leq\|\boldsymbol{W}^{\dagger}\|\|\boldsymbol{C}\|$, we can derive the lower bound of the first-order absolute error for $\boldsymbol{U}^*$,
\begin{equation}
\label{eq_lower}
\frac{1}{\kappa(\boldsymbol{W})} \frac{\|\delta \boldsymbol{C}\|}{\|\boldsymbol{C}\|} \leq \frac{\| \boldsymbol{U}^{*}\|}{\|\delta\boldsymbol{U}^{*}\|}.
\end{equation}
With Eq.(\ref{eq_upper}) and (\ref{eq_lower}), the proof is completed.

\textbf{Remark 1.} From Theorem 3.1, we can note that for a given relative perturbation $\frac{\|\delta \boldsymbol{C}\|}{\|\boldsymbol{C}\|}$, the upper and lower bounds of the relative error over the global field are directly influenced by the condition number. The larger the condition number, the larger the upper bounds and the smaller the lower bounds. This behavior indicates that a large condition number is prone to cause ill-conditioned characteristics of the physical system, where even small noise will lead to significant divergences of results and lose practical significance. To mitigate this, we aim to minimize the condition number, thereby establishing it as a physics-based criterion for sensor placement optimization.

% This indicates that a smaller condition number leads to a smaller error interval and a more stable solution. Therefore, we use the condition number as the physics-based criterion to evaluate sensor location. 
% We note 我们推导得到了重构误差上下界，如果假设受到的相对噪声相同，误差上下界由k决定，条件数越小，上界约小，下界越大，误差区间越小，得到解越稳定。由此我们将此作为传感器指标优化传感器位置。 Thus, the large condition number is prone to case ill-conditioned characteristics of physics system, which means that small noise will cause serious divergence of results and lose practical significance.

\subsection{Validation of proposed method}
% Neural network-based inversion is common practice in machine learning. This powerful learning paradigm is also increasingly used for physical field reconstruction, prediction, and simulations.
Neural network-based modeling is a powerful learning paradigm technique for temperature field reconstruction, prediction, and simulations.
To validate the effectiveness of the proposed method, in this work, we mainly focus on three types of NN-based reconstruction methodologies, non-invasive end-to-end models, non-invasive reduced-order models, and physics-informed models.
% The prevailing reconstruction methodologies that rely on the neural network can be broadly classed into three types: non-invasive end-to-end models, reduced-order models, and physics-informed models.
%画一个图，包含三个模型，3x1
\subsubsection{Non-invasive end-to-end models}
Neural networks, particularly multi-layer perception (MLPs), demonstrate remarkable proficiency in the TFR task, which are non-invasive end-to-end models  \cite{williams2022data, peng2022deep}. The MLP directly learns the mapping relationship from measurements to the temperature field without prior knowledge of physical laws. Fig. \ref{fig:PSPO}(e1) illustrates the conceptual flow of the MLP for the TFR task, and the detailed descriptions are as follows. The MLP can be denoted as
\begin{equation}
\boldsymbol{\hat{u}}_{\boldsymbol{\theta}}(\boldsymbol{s}):=\sigma\left(\boldsymbol{W}^L \sigma\left(\boldsymbol{W}^{L-1}, \cdots, \sigma\left(\boldsymbol{W}^{1} \boldsymbol{s}+\boldsymbol{b}^{1}\right)+\boldsymbol{b}^{L-1}\right)+b^{L}\right),
\end{equation}
where $\boldsymbol{s}$ denotes the measurements, $\sigma$ denotes the nonlinear activation function, and $\boldsymbol{\theta}$ denotes the set containing weights $\boldsymbol{W}^{i}$ and bias $\boldsymbol{b}^{i}$. The training dataset $\mathcal{D}=\left\{\boldsymbol{s}_{i}, \boldsymbol{u}_{i}\right\}_{i=1}^{N}$ consists of pairs of measurements $\boldsymbol{s}_i$ and the corresponding full state of the temperature field $\boldsymbol{u}_i$. With the dataset, we aim to learn a function $\boldsymbol{\hat{u}}_{\boldsymbol{\theta}^*}: \boldsymbol{s} \mapsto \boldsymbol{u}$ from a class of NNs, which can be mathematically expressed as follows,
\begin{equation}
\boldsymbol{\theta}^* \in \underset{\boldsymbol{\theta}}{\arg \min } \frac{1}{N}\sum_{i=1}^N\left\|\boldsymbol{u}_i-\boldsymbol{\hat{u}}_{\theta}\left(\boldsymbol{s}_i\right)\right\|^2.
\end{equation}

% Recently, the deep operator network (DeepONet), another non-invasive end-to-end model, utilized the concept of the universal approximation theorem of operators to overcome the limitation of functional regression, which draws increasing attention to solving high-dimension regression problems. It is an effective deep learning framework to reconstruct the physical field from sparse sensors \cite{lu2021learning, deng2022approximation, goswami2022deep}. 

The DeepONet is a powerful end-to-end NN to solve high-dimension regression problems by combining the universal approximation concept and operator theorem \cite{lu2021learning}. It consists of two sub-networks: a branch net, which encodes the input function at a fixed number of sensors, and a trunk net, which encodes the locations for the output function. Fig. \ref{fig:PSPO}(e2) illustrates the conceptual flow of the DeepONet for the TFR task. 

Concretely, let $\boldsymbol{\hat{u}}$ be an operator that maps an input function $(\boldsymbol{s}, \boldsymbol{\zeta})$ to an output function $\boldsymbol{\hat{u}}_{\boldsymbol{\theta}}(\boldsymbol{s})$, where $\boldsymbol{s}$ denotes the branch input and $\boldsymbol{\zeta} \in \boldsymbol{x}$ denotes points in the domain of the output function. This work uses MLP for the branch and trunk net. 
% The CNN and RNN can be used as the net, while we use the MLP as the baseline model.
The training dataset is $\mathcal{D}=\left\{(\boldsymbol{s}_{i}, \boldsymbol{\zeta}_{i}), \boldsymbol{u}_{i}(\boldsymbol{\zeta}_{i})\right\}_{i=1}^{N}$. The objective of DeepONet is to approximate
\begin{equation}
\boldsymbol{\theta}^* \in \underset{\boldsymbol{\theta}}{\arg \min } \frac{1}{N} \sum_{i=1}^N \left\|\boldsymbol{u}_i(\boldsymbol{\zeta}_{i})-\boldsymbol{\hat{u}}_{\boldsymbol{\theta}}(\boldsymbol{s}_{i})(\boldsymbol{\zeta}_{i}
)\right\|^2.
\end{equation}
% where $w_i$ is the associated weight and is usually considered as unity for every sample in the simplest case.

\subsubsection{Non-invasive reduced-order models}
Proper orthogonal decomposition using NNs \cite{hesthaven2018non, erichson2020shallow, fresca2022pod}, referred to as POD-NN, is a well-established and widely adopted class of reduced-order model techniques. It follows an offline-online paradigm to achieve dimensionality reduction and accurate reconstruction. During the offline stage, POD modes are derived from a set of temperature field snapshots, resulting in a low-dimensional representation of the high-dimensional state space while preserving the essential features of the full-order model. Subsequently, a MLP is utilized to learn the mapping from measurements to POD coefficients. In the online stage, the NN predicts POD coefficients and reconstructed temperature fields are calculated by combining POD modes with coefficients. Fig. \ref{fig:PSPO}(e3) illustrates the conceptual flow of POD-NN for the TFR task. 

Concretely, we have collected $N$ full-sate snapshots $\boldsymbol{u}_i \in \mathbb{R}^n$, which are denoted by the data matrix $\boldsymbol{u}^{t r}=\left[\boldsymbol{u}_1, \boldsymbol{u}_2, \cdots, \boldsymbol{u}_N\right]$. The POD modes are constructed via the singular value decomposition, and the snapshot is estimated with $r$ modes,$
\boldsymbol{u}_i \approx \sum_{j=1}^r \phi_j \alpha_j = \boldsymbol{\Phi}_r \boldsymbol{\alpha}_i$, $\boldsymbol{\Psi}_r \in \mathbb{R}^{n \times r}$ denotes the POD modes of the approximation. $\boldsymbol{\alpha}_i \in \mathbb{R}^{r}$ denotes corresponding POD coefficients and can be calculated by $\boldsymbol{\Phi}_r^\dagger$. The superscript $\dagger$ denotes the Moore-Penrose pseudo-inverse. The measurements $\boldsymbol{s}_{i}$ are obtain with $k$ sensors, $
\boldsymbol{s}_{i}=\boldsymbol{C}_{obs} \boldsymbol{u}_{i} \approx \boldsymbol{C}_{obs} \boldsymbol{\Phi}_r \boldsymbol{\alpha}_{i}
$, $\boldsymbol{C}_{obs}\in \mathbb{R}^{k \times n}$ denotes the matrix determined by sensor locations. In this regard, the training dataset can be established, $\mathcal{D}=\left\{\boldsymbol{s}_{i}, \boldsymbol{\alpha}_{i}\right\}_{i=1}^{N}$. With the training dataset, we aim to learn a function $\boldsymbol{\hat{\alpha}}_{\boldsymbol{\theta}^*}: \boldsymbol{s} \mapsto \boldsymbol{\alpha}$ from a class of NNs, which can be mathematically expressed as follows,
\begin{equation}
\boldsymbol{\theta}^* \in \underset{\boldsymbol{\theta}}{\arg \min } \frac{1}{N}\sum_{i=1}^N\left\|\boldsymbol{\alpha}_i-\boldsymbol{\hat{\alpha}}_{\boldsymbol{\theta}} \left(\boldsymbol{s}_i\right)\right\|^2.
\end{equation}

After training, the NN model can be used to predict new POD expansion coefficients by inputting new measurements. The reconstructed temperature field can be instantly acquired from the linear combination of POD modes. 

\subsubsection{Physics-informed models}
Physics-informed model, particularly physics-informed neural networks (PINN), is a flexible deep learning framework to integrate physical laws with measurements \cite{arzani2021uncovering, liu2022temperature, russell2022physics}. Furthermore, benefiting from physics constraints, the PINN can reduce the requirement of measurements and be easily applied to the TFR task. The conceptual flow of PINN for the TFR task is shown in Fig. \ref{fig:PSPO}(e4), and the detailed description of PINN is provided below.

Let $\boldsymbol{\hat{u}}(\boldsymbol{x})$ be a NN approximating the temperature field $\boldsymbol{u}(\boldsymbol{x})$. The PINN utilizes a MLP architecture. The physics constraints can be given by $\mathcal{L}\boldsymbol{\hat{u}}_{\boldsymbol{\theta}}(\boldsymbol{x})- f(\boldsymbol{x};\boldsymbol{\lambda})=0$ and $\mathcal{B}(\boldsymbol{\hat{u}}_{\boldsymbol{\theta}}(\boldsymbol{x}))=0$. The loss function includes PDE loss, BC loss and data loss to measure the difference between the NN and physics constraints. Therefore, the TFR task in Eq.(\ref{eq_r_task}) can be formulated as 
\begin{equation}
% \mathcal{F} \in \underset{\tilde{\mathcal{F}}}{\arg \min } w_{p d e} \mathcal{L}_{p d e}+w_{b c} \mathcal{L}_{b c}+w_{data} \mathcal{L}_{data},
\boldsymbol{\theta}^* \in \underset{\boldsymbol{\theta}}{\arg \min } \quad w_{p d e} \mathcal{L}_{p d e}+w_{b c} \mathcal{L}_{b c}+w_{data} \mathcal{L}_{data}, \\
\end{equation}
where
\begin{equation}
\begin{aligned}
& \mathcal{L}_{p d e}=\frac{1}{N_f} \sum_{\boldsymbol{x} \in \mathcal{P}_f}\|\mathcal{L}\boldsymbol{\hat{u}}_{\boldsymbol{\theta}}(\boldsymbol{x}) - f(\boldsymbol{x};\boldsymbol{\lambda})\|^2, \\
& \mathcal{L}_{b c}=\frac{1}{N_{b c}} \sum_{\boldsymbol{x} \in \mathcal{P}_{b c}}\left\|\mathcal{B}(\boldsymbol{\hat{u}}_{\boldsymbol{\theta}}(\boldsymbol{x}))\right\|^2, \\
& \mathcal{L}_{data}=\frac{1}{N_{o b s}} \sum_{\boldsymbol{x}_{loc} \in \mathcal{P}_{data }}\|\boldsymbol{s}-\boldsymbol{\hat{u}}_{\boldsymbol{\theta}}(\boldsymbol{x}_{loc})\|^2,
\end{aligned}
\end{equation}
and $w_{pde}$, $w_{bc}$ as well as $w_{data}$ are predefined weight hyper-parameters. The relative magnitude of the loss terms significantly impacts the model training and prediction.
% The relative magnitude of the loss terms will affect the model training and results. If the weight of the loss term, or its derivative relative to the model is significantly greater than that of other losses, the neural network will train to minimize the loss term. 
Therefore, the hyper-parameters are set to balance relative magnitude of multiple loss terms \cite{liu2022temperature}. $\mathcal{P}_{pde}$ and $\mathcal{P}_{bc}$ denote the sets of residual points from PDE and BC, which are obtained through sampling randomly in the domain $\Omega$.  $\mathcal{P}_{data}$ denote the sets containing measurements.

% A typical sensor placement method is primarily characterized by the objective function (performance criterion) and the optimization technique (selection procedure).

% We propose a novel sensor placement method using signal reconstruction error as the cost function, sequentially minimize it with greedy procedures.

% we propose a novel sensor placement method using signal reconstruction error as the cost function, sequentially minimize it with greedy procedures.% The

% The sensor placement in Eq.(\ref{optima_problem}) is a combinatorially hard problem 

\section{Numerical result}
\label{sec result}
In our problem setting, there is insufficient data to construct data-based criteria for sensor placements. Therefore, we propose a novel physics-driven sensor placement optimization (PSPO) method in data-free scenarios. To validate the effectiveness of the PSPO method, we consider sampling sensor locations by the uniform sampling technique (US) and the random technique (RS) as the baseline. Furthermore, assuming sufficient data to construct data-based criteria, we further compare the PSPO method with data-driven sensor placements optimization methods, including ConditionNumberSampler (CNS) \cite{jayaraman2019interplay, manohar2018data, de2021pysensors} and EnhancedClusteringSampler (ECS) \cite{dubois2022machine}. On the one hand, reduced-order model considers reconstruction problems as a linear least squares problem through the linear superposition of POD modes. CNS uses the condition number as a criterion for error sensitivity and equation stability, specifically the condition number of the product of the measurement matrix and a tailored library extracted from data. Greedy algorithm is then utilized to sequentially select locations with the smallest condition numbers, determining the best placement. On the other hand, ECS uses Voronoi tessellation to partition the computational area, preventing sensor clustering. The centroid of each Voronoi polygon is extracted, and correlation coefficients based on the data are calculated for clustering. Sensors are placed at the least correlated locations to determine the best placement.
% using the singular value decomposition and QR pivoting. 

% On the premise that large measured data cannot be obtained, such as the experimental design stage, data-driven sensor placement methods are not practical. However, to investigate the effectiveness of the proposed method,
% we assume there is sufficient data to construct data-based criteria and further compare the PSPO method with data-driven sensor placements optimization methods, including CNS \cite{manohar2018data, de2021pysensors} and ECS \cite{dubois2022machine}.

% because the accuracy archived by the greedy method is similar to other existing methods and the computational cost is smaller than the other methods as shown in Ref

\subsection{1D heat equation}
A simple one-dimensional (1D) case is considered to demonstrate and analyze our proposed method. The 1D steady-state heat equation is described as 
\begin{equation}
\label{eq:case1_pde}
u_{x x}=f(x;\boldsymbol{\lambda}), x \in \Omega,
\end{equation}
where $u$ denotes the temperature field, $\Omega=\{x|x \in[-10, 10]\}$, and the source term $f(x;\boldsymbol{\lambda})$ is composed of two wave functions, $f(x;\boldsymbol{\lambda})=-\lambda_1 \cdot \sin (0.7 x)-\lambda_2 \cdot \cos (1.5 x)$. The intensities of sources are unknown problem parameters and $\lambda_1, \lambda _2 \in [0, 20]$. The boundary values are determined by the following exact solution,  
\begin{equation}
\label{eq:case1_exact}
u=\frac{\lambda_1}{0.49} \sin (0.7 x)+\frac{\lambda_2}{2.25} \cos (1.5 x)-0.1 x.
\end{equation}
Assuming the mathematical model is partially known (problem parameters are unknown), we only have $k$ sensors in $\Omega$. Our TFR task is to reconstruct the global temperature field using appropriate sensor placement. More details refer to \cite{liu2022novel}.

We first select sensor locations by our PSPO method. By default, the number of point stencils and nearest neighbor points are $400$ and $25$. The 3rd-order polyharmonic splines as the radial basis function. In GA, the population size is set to $10$, and the maximum number of generations is set to $2,000$. The crossover probability and mutation probability are set to $0.9$ and $1/k$, respectively. $k$ denotes the sensor number. The fitness evaluation function is the physics-based criterion. Then we use MLP, DeepOnet, PODNN, and PINN models with the best sensor locations to reconstruct the temperature field. The $80$ training and $20$ test data are generated by randomly sampling from problem parameters for MLP, DeepOnet, and PODNN. As for PINN, we use the measurements under $\lambda_1=0.49, \lambda_2=2.25$. $w_{pde}$, $w_{bc}$ and $w_{data}$ are set to be $1$ to balance relative magnitude of multiple loss terms. The configurations for reconstruction models are shown in Table \ref{tab:case1_nn}, where $k$ denotes the sensor number and POD modes are set to $10$. The baseline settings of neural network are determined according to different experiments and related literature references \cite{erichson2020shallow, lu2021learning, liu2022temperature}. The gradient-based optimization solver is Adam. The activation function is GELU, and the learning rate is $0.001$. Especially, DeepOnet uses two sub-networks to map  the input function including measurements and points in the domain of the output function to the reconstructed temperature field, which requires to extract the feature. Besides, DeepOnet can not achieve acceptable predictions when the neuron number is set to be $64$ in experiments. Therefore, DeepOnet has twice as many neurons as other models in this work.

\begin{table*}[!t]
\centering
\caption{Configurations for reconstruction models.}
\scalebox{0.86}{
\begin{tabular}{ccc}
\hline Model& Layers& Adam Iterations \\ \hline
MLP& {[}k, 64, 64, 64,400{]}& 4000 \\
\multirow{2}{*}{DeepOnet} & Trunk net:{[}k, 128, 128, 128,128{]} & \multirow{2}{*}{10000} \\ & Branch net: {[}1,128, 128,128,128{]} &\\PODNN & {[}k, 64, 64, 64,10{]} & 4000  \\
PINN& {[}1, 64, 64, 64,1{]}  & 4000 \\ \hline
\end{tabular}
}
\label{tab:case1_nn}
\end{table*}

\begin{figure*}[!b]
	\centering
\includegraphics[width=1\linewidth]{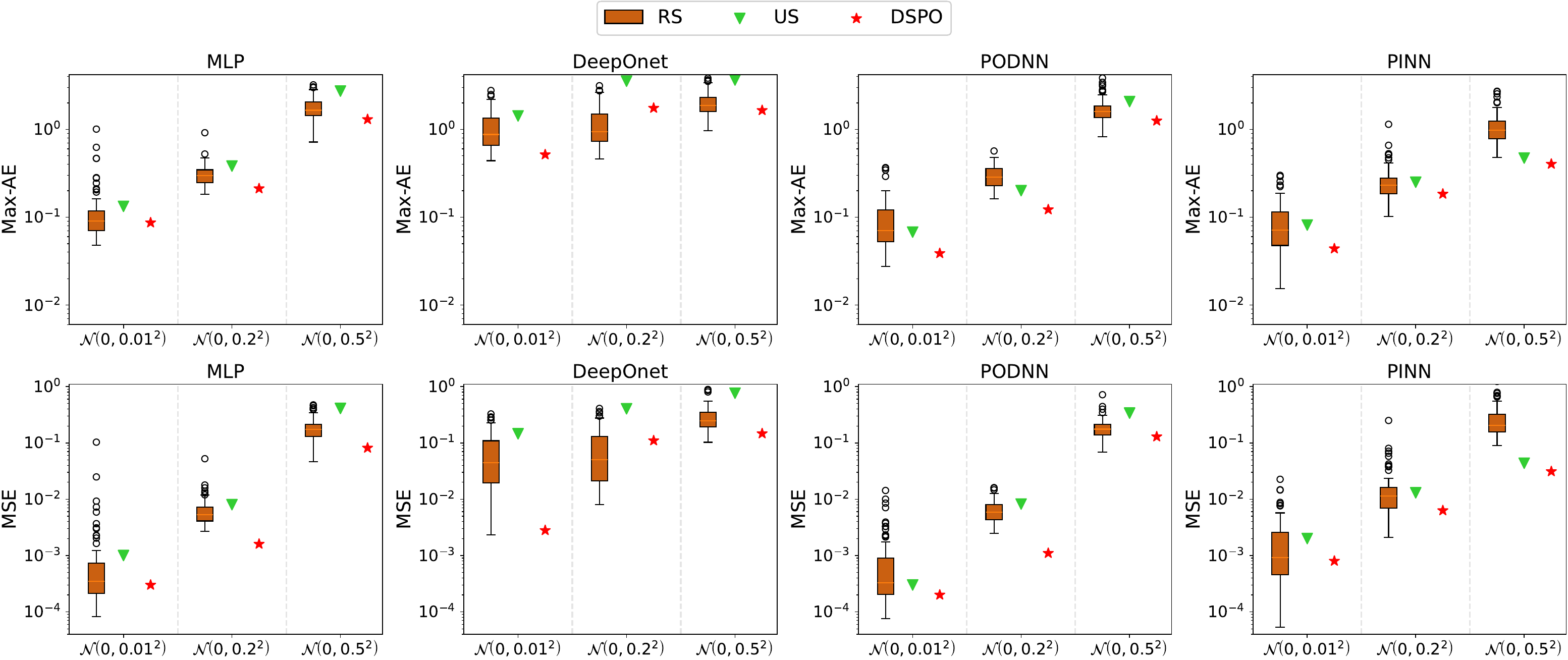}
	\caption{1D heat equation: comparisons of MLP, DeepOnet, PODNN, and PINN using PSPO, RS, and US under three noise scales. The first two rows represent Max-AEs and MSEs. The columns (from left to right) represent the results of MLP, DeepOnet, PODNN, and PINN, respectively. The black circle represents an outlier in a box plot. Error bar (square) represents results using 100 strategies of sensor placement randomly generated by RS.}
	\label{fig:case1_noise}
\end{figure*}

% \vspace{2cm}

Fig. \ref{fig:case1_noise} illustrates the reconstruction performances of four models using PSPO, US, and RS ($100$ sets of sensor locations generated randomly). The noise level $\epsilon$ is chose from from $\{1 \%, 10 \%, 50 \%\}$ and the sensor number is equal to $10$. Maximum Absolute Error (Max-AE) and Mean Squared Error (MSE) are calculated to evaluate reconstruction performances. Error bar (square) illustrates results using 100 strategies of sensor placement randomly generated by RS. Triangle and star represent results of US and DSPO, respectively. For all four reconstruction models under three noise levels, Max-AEs and MSEs by PSPO are much smaller than that by US. Moreover, Max-AEs and MSEs by PSPO are superior to the medians by RS except for DeepOnet under $\varepsilon_u \sim \mathcal{N}\left(0,0.1^2\right)$. It is important to note that RS exhibits large variances and is prone to lead to outlier results. This indicates RS can not guarantee a relatively promising result. In conclusion, when there is insufficient data for data-based criteria, PSPO can provide best sensor locations to improve performances for reconstruction models. Besides, Max-AEs and MAEs of the four models tend to increase with the noise level increase. In terms of reconstruction performance, PODNN outperforms the other models, while DeepOnet is the worst. It is worth noting that DeepOnet exhibits large errors even under low noise conditions, which can be attributed to the inherent limitations in its neural architecture for reconstruction tasks. DeepOnet encodes coordinates of output query points to predict the field at continuous locations, which increases learning complexity and leads to larger errors. Consequently, its reconstruction performance is inferior to other models like MLP, PODNN, and PINN. Despite its slightly lower reconstruction performance, the primary advantage of DeepOnet is its ability to output continuous values of the global field at arbitrary locations, unlike other reconstruction methods that provide discrete values at fixed coordinates. More importantly, DeepOnet with DSPO outperforms DeepOnet with RS and US, indicating excellent compatibility between DeepOnet and sensor placement. This flexibility and compatibility can lead to improved performance. While DeepOnet may have relatively higher errors in certain conditions, its unique strengths make it valuable in specific applications. Additionally, Table \ref{tab:case1_time} presents comparisons of objective function and computational cost between PSPO, RS, and US. Under the different sensor numbers, PSPO achieves better objective function values compared to RS and US. The experiments with different neural network configurations are summarized in Table \ref{tab:net_case1}. Among all models, the baseline model consistently demonstrated superior performance, in terms of MSEs. Therefore, the baseline setting is used for the following experiments.

\begin{table}[!ttp]
\centering
\caption{1D heat equation: comparisons of RS, US and DSPO in terms of condition number and computational cost. The best results under the different sensor number are highlight.}
\scalebox{0.86}{
\begin{tabular}{cccccccccccc}
\toprule
% \multirow{2}{*}{Method} & \multicolumn{11}{c}{Sensors}                                                                                \\ \cline{2-3} \cline{5-6} \cline{8-9} \cline{11-12} 
\multirow{2}{*}{Method}
                        & \multicolumn{2}{c}{5}  &  & \multicolumn{2}{c}{15} &  & \multicolumn{2}{c}{20} &  & \multicolumn{2}{c}{25} \\
                          \cline{2-3} \cline{5-6} \cline{8-9} \cline{11-12}
                        & $log(\kappa(\boldsymbol{W}))$ & Time/min     &  & $log(\kappa(\boldsymbol{W}))$ & Time/min     &  & $log(\kappa(\boldsymbol{W}))$ & Time/min     &  & $log(\kappa(\boldsymbol{W}))$ & Time/min     \\ \hline
RS                      & 10.94   & \textless{}1 &  & 10.18   & \textless{}1 &  & 10.15   & \textless{}1 &  & 10.03   & \textless{}1 \\
US                      & 11.43   & \textless{}1 &  & 10.63   & \textless{}1 &  & 10.42   & \textless{}1 &  & 10.28   & \textless{}1 \\ \hline
DSPO          & \textbf{10.78}           & 5.60        &  & 10.07   & 6.31        &  & \textbf{9.92}  & 6.80        &  & \textbf{9.82}    & 7.01       \\ \toprule
\end{tabular}
}
\label{tab:case1_time}
\end{table}

\begin{table}[!ttp]
\centering
\caption{1D heat equation: comparisons of four models with different parameters setting. The sensor locations are obtained through DSPO. The noise level is $\varepsilon_u \sim \mathcal{N}\left(0,0.1^2\right)$. The number of observations is $5$. The best results under the different sensor number are highlight.}
\scalebox{0.86}{
\begin{tabular}{ccccc}
\toprule
\multicolumn{2}{c}{Model}            & Layers                                                                                                        & Max-AE         & MSE            \\ \hline
\multirow{3}{*}{MLP}      & Baseline & {[}5,64,64,64,400{]}                                                                                          & \textbf{0.560} & \textbf{0.028} \\
                          & Deepen   & {[}5,64,64,64,64,400{]}                                                                                       & 1.678          & 0.2046         \\
                          & Widen    & {[}5,128,128,128,400{]}                                                                                       & 0.567          & 0.032          \\ \hline

\multirow{3}{*}{PODNN}    & Baseline & {[}5,64,64,64,10{]}                                                                                           & \textbf{0.268} & \textbf{0.007} \\
                          & Deepen   & {[}5,64,64,64,64,10{]}                                                                                        & 0.299          & 0.008          \\
                          & Widen    & {[}5,128,128,128,10{]}                                                                                        & 0.269          & \textbf{0.007} \\ \hline
\multirow{3}{*}{PINN}     & Baseline & {[}1,64,64,64,1{]}                                                                                            & 0.597          & \textbf{0.067} \\
                          & Deepen   & {[}1,64,64,64,64,1{]}                                                                                         & \textbf{0.554} & 0.083          \\
                          & Widen    & {[}1,128,128,128,1{]}                                                                                         & 0.660          & 0.076          \\ \hline
\multirow{4}{*}{DeepOnet} & Baseline & \begin{tabular}[c]{@{}c@{}}Net1:{[}5, 128,128,128,128{]}\\ Net2:{[}1, 128,128,128,128{]}\end{tabular}         & 2.523          & \textbf{0.207} \\
                          & Deepen   & \begin{tabular}[c]{@{}c@{}}Net1:{[}5, 128,128,128,128,128{]}\\ Net2:{[}1, 128,128,128,128,128{]}\end{tabular} & \textbf{2.284} & 0.348          \\
                          & Widen    & \begin{tabular}[c]{@{}c@{}}Net1:{[}5, 256,256,256,256{]}\\ Net2:{[}1, 256,256,256,256{]}\end{tabular}         & 3.365          & 0.226          \\ \toprule
\end{tabular}
}
\label{tab:net_case1}
\end{table}

\begin{figure*}[!t]
	\centering
\includegraphics[width=1.0\linewidth]{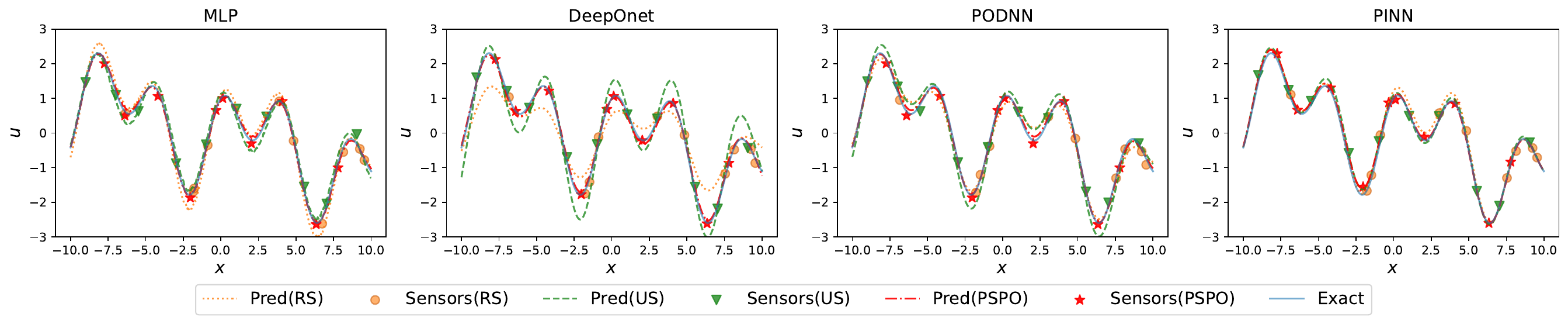}
	\caption{1D heat equation: predictions of MLP, DeepOnet, PODNN, and PINN using PSPO, RS, and US under $\lambda_1=0.49$, $\lambda_2=2.25$, $\varepsilon_u \sim \mathcal{N}\left(0,0.1^2\right)$ and $k=10$.}
	\label{fig:case1_pred}
\end{figure*}

Fig. \ref{fig:case1_pred} shows a set of predictions using PSPO, RS, and US under $\lambda_1=0.49$, $\lambda_2=2.25$, $\varepsilon_u \sim \mathcal{N}\left(0,0.1^2\right)$ and $k=10$. For MLP, DeepOnet, PODNN, and PINN, the predictions by PSPO are much closer to the exact values than that by RS and US. In particular, compared with RS and US, the reconstruction performance by PSPO is improved more significantly for DeepOnet. Furthermore, sensor locations by PSPO are near peaks and troughs of the exact curve. In contrast, RS and US ignore the peaks and troughs due to randomness and uniformity of sampling. The exact solution in Eq.(\ref{eq:case1_exact}) follows the wave function. The peaks and troughs are the critical locations where the wave function undergoes the most significant changes. These regions provide important information about the amplitude and phase of the wave function, which is essential for achieving higher precision reconstruction. In a word, PSPO effectively captures significant change locations by combining with the physics information in Eq.(\ref{eq:case1_pde}) to improve the reconstruction performance.

% \vspace{1.5cm}

\begin{table*}[!tbp]
\centering
\caption{1D heat equation: comparisons of four models by PSPO, US, CNS and ECS under $k=3, 10, 20$. The noise level is $\varepsilon_u \sim \mathcal{N}\left(0,0.1^2\right)$. The best results under the different sensor number are highlight.}
\scalebox{0.86}{
\begin{tabular}{cccccccccc}
\toprule
\multirow{2}{*}{Model}    & \multirow{2}{*}{Method}   & \multicolumn{2}{c}{3}   & \multicolumn{2}{c}{10}                      & \multicolumn{2}{c}{20}                      & \multicolumn{2}{c}{best/all}  \\ \cline{3-10} &  & Max-AE   & MSE      & Max-AE  & MSE  & Max-AE & MSE & Max-AE & MSE  \\ \hline
\multirow{4}{*}{MLP}      & PSPO  & \textbf{0.7631}      & 0.0315               & \textbf{0.2118}      & 0.0030               & \textbf{0.1022}      & 0.0019               & \textbf{3/3}         & 0/3\\ & US            & 1.0346               & 0.0917 & 0.4820               & 0.0080               & 0.3880               & 0.0044               & 0/3                  & 0/3  \\ & CNS    & 1.0168  & 0.0234  & 0.3058               & \textbf{0.0023}      & 0.2756               & \textbf{0.0010}      & 0/3                  & \textbf{2/3}         \\ & ECS & 1.0224               & \textbf{0.1515}      & 0.7265               & 0.0071 & 0.4790               & 0.0027 & 0/3                  & 1/3                  \\ \hline
\multirow{4}{*}{DeepOnet} & PSPO                      & 3.4371               & 0.6923               & \textbf{1.7469}      & \textbf{0.1099}      & \textbf{0.4283}      & \textbf{0.0101}      & \textbf{2/3}         & \textbf{2/3}         \\ & US  & 4.9640 & 1.3618               & 3.5337               & 0.4032               & 1.2238               & 0.2278               & 0/3   & 0/3\\ & CNS    & \textbf{3.0810}      & \textbf{0.4550}      & 1.9269               & 0.1249               & 0.6330               & 0.0120               & 1/3                  & 1/3  \\ & ECS & 3.7282 & 0.5760 & 2.7549               & 0.1513  & 0.6930               & 0.0339               & 0/3    & 0/3  \\ \hline\multirow{4}{*}{PODNN}    & PSPO  & 0.6101               & 0.0184               & \textbf{0.1221}      & \textbf{0.0011}      & \textbf{0.1215}      & 0.0002               & \textbf{2/3}         & 1/3                  \\& US            & 0.7620               & 0.0491               & 0.1707               & 0.0082               & 0.1257               & 0.0036  & 0/3                  & 0/3                  \\& CNS    & \textbf{0.5711}      & 0.0216       & 0.1335               & 0.0016               & 0.1348     & 0.0003               & 1/3                  & 0/3      \\& ECS & 0.6032               & \textbf{0.0183}      & 0.3083               & 0.0018               & 0.1360               & \textbf{0.0001}      & 0/3                  & \textbf{2/3}         \\ \hline
\multirow{4}{*}{PINN}     & PSPO                      & \textbf{0.9780}      & \textbf{0.1348}      & 0.1840               & 0.0063               & \textbf{0.1141}      & 0.0046               & \textbf{2/3}         & \textbf{1/3}         \\& US            & 2.4690  & 1.0630               & 0.2504               & 0.0130               & 0.1855               & 0.0056 & 0/3       & 0/3  \\ & CNS    & 1.9380  & 1.0340   & \textbf{0.1503} & 0.0051               & 0.1554  & \textbf{0.0024}      & 1/3  & \textbf{1/3}         \\& ECS & 3.7260 & 1.0600               &0.3080   & \textbf{0.0105} & 0.1345&0.0037 & 0/3  & \textbf{1/3}         \\ \toprule
\end{tabular}
}
\label{tab:case1_num}
% \vspace{-40pt}
\end{table*}

\begin{figure*}[!t]
	\centering
\includegraphics[width=1.0 \linewidth]{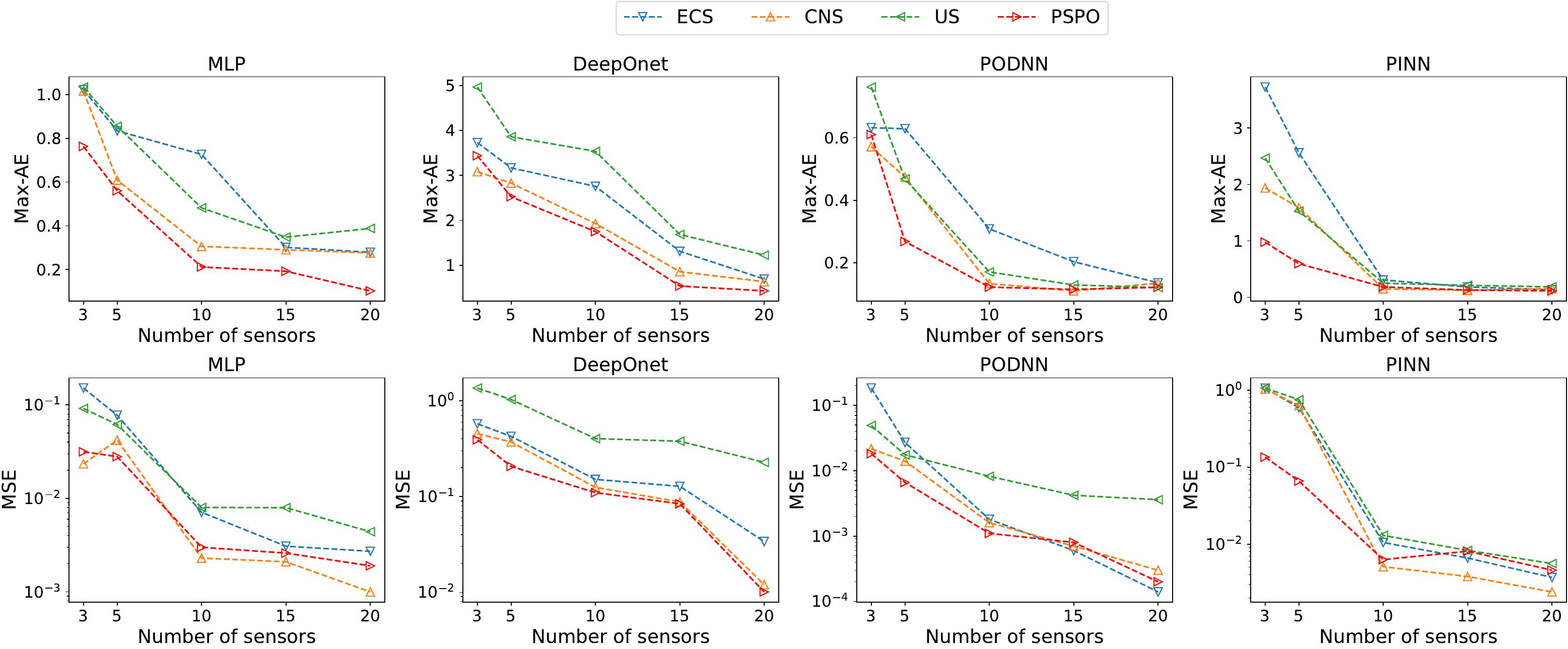}
	\caption{1D heat equation: comparisons of MLP, DeepOnet, PODNN, and PINN with PSPO, US, CNS, and ECS under the different sensor number. The row represents MAx-AEs and MSEs. The columns (from left to right) represent results of MLP, DeepOnet, PODNN, and PINN, respectively.}
	\label{fig:case1_num}
\end{figure*}

To further investigate the effectiveness of PSPO, we assume there is sufficient data available for CNS and ECS. Table \ref{tab:case1_num} presents comparisons between PSPO, US, CNS, and ECS under the different sensor numbers. The Max-AEs and MSEs by PSPO are better than that by US. Moreover, models by PSPO tend to achieve the best Max-AE. For MSEs, PSPO is always more than or close to the best performance achieved by CNS or ECS. More importantly, while CNS and ECS with sufficient datasets directly utilize extensive information from global fields for sensor selection, our proposed PSPO method operates under stricter constraints, performing sensor selection without relying on any dataset. This capability allows the PSPO method to offer sensor placement strategies in data-free scenarios, a significant advantage that CNS and ECS cannot easily achieve. Therefore, even if the PSPO method shows slightly inferior performance compared to CNS and ECS under data-rich conditions, its ability to function effectively without data makes it a valuable alternative. It is more intuitive to get the above conclusion from Fig.\ref{fig:case1_num}. As the figure shows, PSPO is superior to US, and the trends of Max-AEs and MAEs by PSPO, CNS, and ECS are close. This implies that models by PSPO can gain a similar reconstruction performance as models by data-driven sensor placement optimization methods.
% \usepackage{multirow}
% \usepackage{multirow}

\begin{figure*}[!b]
	\centering
\includegraphics[width=0.85 \linewidth]{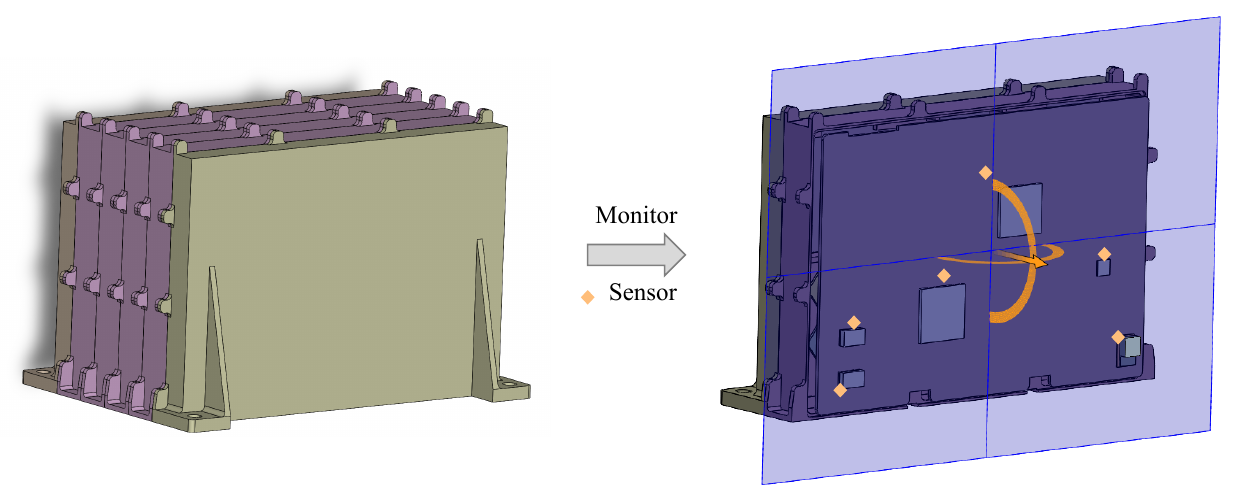}
	\caption{Application case: a signal processing module. Sensors are placed over the process plate to monitor the global temperature field.}
	\label{fig:case2_problem}
\end{figure*}

\subsection{Application: a signal processing module}
The device is three dimensional signal processing module with a stepped structure, shown in Fig. \ref{fig:case2_problem}. This device consisting of six plates, including a power circuit plate, a conditioning circuit plate, two identical signal processing plates, and front and rear cover plates. However, in this practical application, the power and conditioning plates generate minimal heat, so temperature monitoring is generally unnecessary for these plates. However, signal processing plates include key digital processing chips, such as DSP signal processors, LTM4644EY regulators, FPGA chips, and analog-to-digital converters, all of which have high thermal power density and require focused temperature monitoring. Therefore, in this application, we require to perceive the global two dimensional temperature field on the processing plate for health monitoring and active thermal control, ultimately improving lifespan and reliability.

The two dimensional signal processing plate, which has six source components and limited sensors are placed over the plate. The maximum number of sensors is limited to 20. Our proposed DSPO method is used to improve reconstruct accuracy through optimizing sensor placement at a given number of sensors. The plate is considered as an application case to demonstrate and analyze our proposed DSPO method. Concretely, the steady-state temperature field $T$ of the process plate satisfy
\begin{equation}
\label{case2}
\begin{array}{r}
\frac{\partial}{\partial x}\left(k \frac{\partial T}{\partial x}\right)+\frac{\partial}{\partial y}\left(k \frac{\partial T}{\partial y}\right)+\phi(x, y)=0, \vspace{1ex}\\
Boundary:  \quad T=T_{0} \quad  or  \quad k \frac{\partial T}{\partial \mathbf{n}}=0,
\end{array}
\end{equation}
where $k$ denotes the thermal conductivity, $\phi$ denotes the intensity distribution function, and $T_{0}$ denotes the temperature value of the isothermal boundary or the convective boundary conditions. Eq.(\ref{case2}) represents three boundary conditions, including Dirichlet (isothermal) and Neumann (adiabatic) boundary conditions. The area of a single piece of heat source component is set as a rectangle, modeled as
\begin{equation}
\label{eq3}
\phi(x, y)=\left\{\begin{array}{ll}
\phi_{i}, & (x, y) \in \Gamma_{i}, i=1,2, \cdots, n_s \\
0, & (x, y) \notin \Gamma_{i}
\end{array}\right.,
\end{equation}
where $\phi_{i}$ denotes the (constant) intensity of $i$th heat source, $n_s$ denotes the number of heat sources and $\Gamma_{i}$ denotes the area of $i$th heat source. The intensities of sources are the problem parameters, which are unavailable when components work. Assuming the mathematical model is partially known (the problem parameters are unknown), we only have $k$ sensors in the $\Omega$. Our task is to reconstruct the global temperature field using appropriate sensor placement. The Patran software, based on finite element method, is used for the heat transfer analysis of the processing module. The modeling process is as follows. First, the three-dimensional solid model is simplified by omitting components with lower heat dissipation, such as minor components, circuit connections, and sensor wires. Next, the simplified model is meshed and use unstructured grids. Finally, the model parameters are set, including the material properties of the outer frame, circuit plates, and chips. Thermal contact conditions are established, the heat source powers of chips is defined, and boundary conditions are specified. This numerical model comprises 199,896 element and use unstructured discretization schemes with tetrahedral mesh. The model is configured with the bottom surface held at a constant temperature of $10\,^{\circ}\mathrm{C}$. The remaining boundaries simulate adiabatic conditions in a vacuum. The initial temperature within the model is uniformly set to $0\,^{\circ}\mathrm{C}$. Our reconstruct dataset use the temperature data of the process plate, which has 1,175 grid points. The thermal conductivity of the plate is $150 \text{W/m}\cdot\text{K}$. The mesh independence comparisons of of maximum, minimum, and average temperatures on the plate are shown in the Table \ref{tab:case2_mesh}. The maximum, minimum, and average temperatures are close under different element number, indicating that the numerical simulation results have converged. Thus, the mesh independence verification is complete.

\begin{table}[!h]
\centering
\label{tab:case2_nn}
\caption{Mesh independence comparison of maximum temperature, minimum temperature, and average temperature.}
\begin{tabular}{cccc}
\hline
Element Number & Maximum T & Minimum T & Average T \\ \hline
183,900     & 71.439    & 19.0810   & 32.468    \\
199,896     & 71.431    & 19.074    & 32.467    \\
219,884     & 71.426    & 19.071    & 32.466    \\ \hline
\end{tabular}
\label{tab:case2_mesh}
\end{table}
% \vspace{2cm}

\vspace{7cm}
The PSPO method is used to select sensor locations. By default, the number of point stencils and nearest neighbor points are $1370$ and $25$. Radial basis functions are 3rd-order polyharmonic splines. In GA, the population size is $20$, and the maximum number of iterations is $10,000$. The crossover probability and mutation probability are $0.9$ and $1/k$, respectively. $k$ denotes the sensor number. The fitness evaluation function is the physics-based criterion. Then we reconstruct the temperature field using MLP, DeepOnet, PODNN, and PINN. The $1600$ training data and $400$ test data are generated by randomly sampling from problem parameters $\phi_{i}$. For PINN, $w_{pde}$, $w_{bc}$ as well as $w_{data}$ are set to be $1$, $1$ and $100$ to balance relative magnitude of multiple loss terms. The configurations for reconstruction models are shown in Table \ref{tab:case2_nn}, where $k$ denotes the sensor number and POD modes are set to $20$. The baseline settings of neural network are determined according to different experiments and related literature reference \cite{erichson2020shallow, lu2021learning, liu2022temperature}. The gradient-based optimization solver is Adam. The activation function is GELU, and the learning rate is $0.001$.

\begin{table}[!t]
\centering
\caption{Configurations for reconstruction models.}
\begin{tabular}{ccc}
\toprule
Model& Layers& Iterations \\ \hline
MLP& {[}$k$, 64, 64, 64, 1175{]}& 10000 \\
\multirow{2}{*}{DeepOnet} & Trunk net:{[}$k$, 128, 128, 128,128{]} & \multirow{2}{*}{20000} \\ & Branch net: {[}2,128, 128,128,128{]} &\\PODNN & {[}$k$, 64, 64, 64, 20{]} & 10000  \\
PINN& {[}2, 64, 64, 64, 1{]}  & 10000 \\ \toprule
\end{tabular}
\label{tab:case2_nn}

\end{table}

\subsubsection{Performance with different noise level}
This experiment is designed to validate the effectiveness of PSPO under various noise levels. The sensor number is set to be $10$, and the noise level $\epsilon$ is chosen from from $\{1 \%, 10 \%, 50 \%\}$. The commonly used noise level $\epsilon$ is $10\%$. Additionally, to explore the effectiveness of the PSPO method, we considered both low-noise and high-noise scenarios, represented by $1\%$ and  $50\%$ noise levels, respectively. This setting allows us to evaluate the robustness and performance of PSPO under varying noise levels. RS and US are used as the baseline, whereas RS is used for $200$ independent trials to obtain sensor locations.

\begin{figure*}[!h]
\centering
\includegraphics[width=1.0 \linewidth]{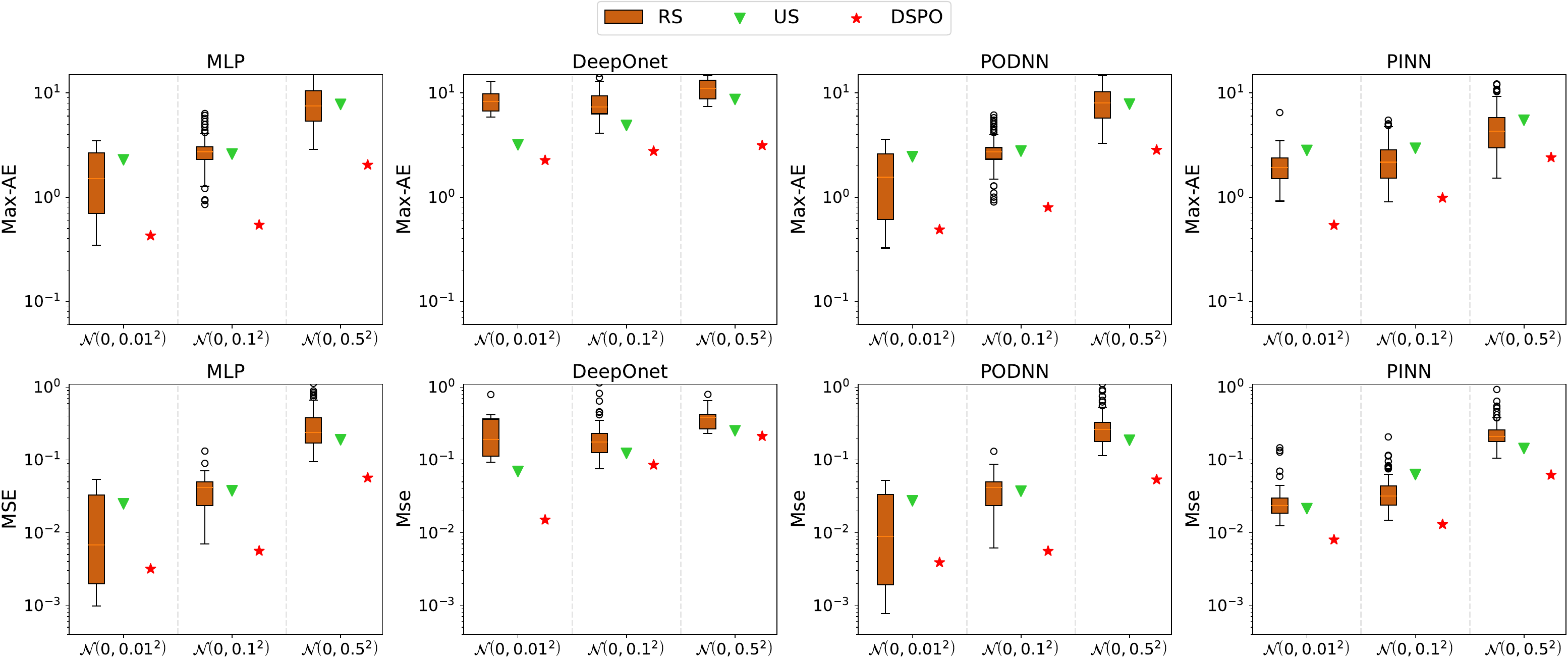}
\caption{Application case: comparisons of MLP, DeepOnet, PODNN, and PINN with PSPO, RS, and US under $\varepsilon_u \sim \mathcal{N}\left(0,0.01^2\right)$, $\varepsilon_u \sim \mathcal{N}\left(0,0.1^2\right)$, and $\varepsilon_u \sim \mathcal{N}\left(0,0.5^2\right)$. The first two rows represent Max-AEs and MSEs. The columns (from left to right) represent the results of MLP, DeepOnet, PODNN, and PINN, respectively. The black circle represents an outlier in a box plot.}
\label{fig:case2_noise}
\end{figure*}

% \vspace{2cm}
Fig. \ref{fig:case2_noise} shows the reconstruction performances of four models using PSPO, US, and RS. As the noise level increases, Max-AEs and MSEs continue to increase. For the four reconstruction models under three noise levels, Max-AEs and MSEs by PSPO are smaller than that by US and the median by RS. For DeepOnet and PINN, PSPO outperforms the best result by RS. Moreover, since RS strongly depends on the random seed used, it has large variances and is prone to cause singular values. Some singular values are significantly higher than the average value by more than $50\%$, which are much worse than US and PSPO.
Furthermore, in terms of reconstruction performance, DeepOnet performs worse than the other three models. Interestingly, MLP and PODNN exhibit similar reconstruction performances. In a word, PSPO is a powerful method to solve sensor placement optimization problems in data-free scenarios, providing better sensor locations for models to achieve better performance.

% PSPO provides optimized sensor locations for models to achieve better performance data-free scenarios.

\begin{figure*}[!t]
	\centering
\includegraphics[width=1.0 \linewidth]{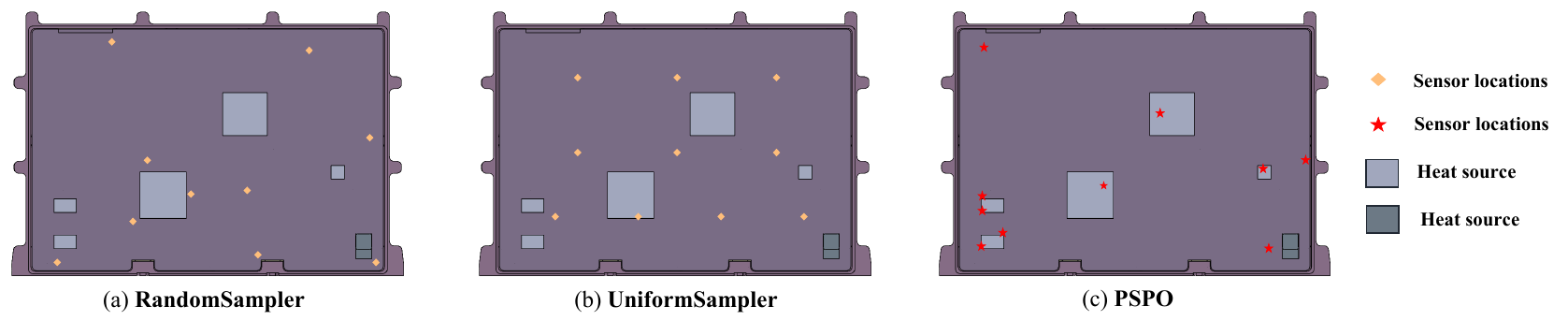}
	\caption{Application case: sensor locations generated by PSPO, RS, and US under $k=10$.}
	\label{fig:case2_sensor_compare}
\end{figure*}

\begin{figure*}[!h]
	\centering
\includegraphics[width=1.0 \linewidth]{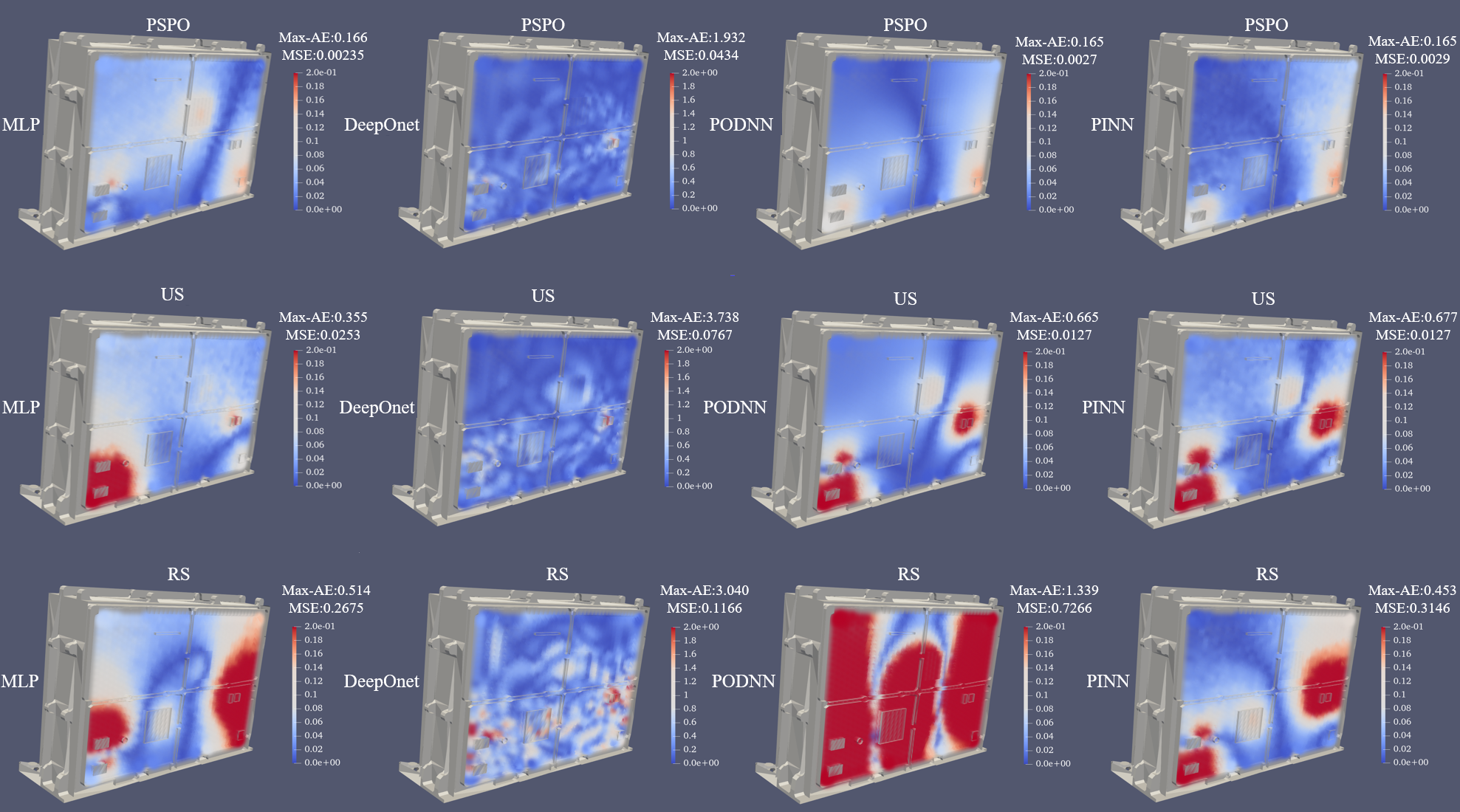}
	\caption{Application case: point-wise errors of MLP, DeepOnet, PODNN, and PINN using PSPO, RS, and US under $k=10$ and $\epsilon_{u} \sim \mathcal{N}\left(0, 0.01^{2}\right)$. The columns (from left to right) represent the results of MLP, DeepOnet, PODNN, and PINN, respectively. The rows represent results by PSPO, RS, and US, respectively. Besides, it is worth noting that the maximum error interval is $2$ for DeepOnet and $0.2$ for other models.}
	\label{fig:case2_pred_compare}
\end{figure*}

Fig. \ref{fig:case2_sensor_compare} shows sensor locations generated by RS, US, and PSPO under $k=10$. As the figure shows, most sensors by PSPO are close to heat sources, and a few are near the boundary, which provides important information about heat source and boundary. This is because heat sources are the main source of the temperature field, and the temperature change is most apparent around the heat source. In addition, sensors near the heat source can also provide information about the heat source, such as locations and sizes of heat sources, which is very important for learning the characteristics of the global temperature field distribution. However, RS and US consider randomness and uniformity in space, ignoring the physical information of temperature fields. Fig. \ref{fig:case2_pred_compare} shows a set of point-wise errors by RS, US, and PSPO, respectively. The point-wise error by PSPO is much less than RS and US.

% \subsubsection{Performance with different noise types}

\begin{table*}[!h]
\centering
\caption{Application case: comparisons of four models using PSPO, US, CNS and ECS under $k=6, 16, 36$, and $\varepsilon_u \sim \mathcal{N}\left(0,0.1^2\right)$. The best results under the different sensor numbers are highlighted.}
	\scalebox{0.86}{
\begin{tabular}{cccccccccc}
\toprule
\multirow{2}{*}{Model}    & \multirow{2}{*}{Method}   & \multicolumn{2}{c}{6} & \multicolumn{2}{c}{16}& \multicolumn{2}{c}{36} & \multicolumn{2}{c}{best/all} \\ \cline{3-10}& & Max-AE          & MSE             & Max-AE          & MSE             & Max-AE          & MSE             & Max-AE        & MSE       \\ \hline
\multirow{4}{*}{MLP}      & PSPO                      & 1.5924          & 0.0337          & \textbf{0.5075} & 0.0033          & \textbf{0.4092} & \textbf{0.0037}          & \textbf{2/3}           & \textbf{1/3}          \\ & US            & 2.7371          & 0.0517          & 2.4430          & 0.0176          & 1.0970          & 0.0044          & 0/3           & 0/3          \\ & CNS    & \textbf{0.8510} & 0.0083          & 0.6436          & \textbf{0.0032} & 0.5630          & 0.0038 & 0/3           & \textbf{1/3}          \\& ECS & 2.0953          & \textbf{0.0251} & 0.6377  & 0.0036          & 0.5270          & 0.0041          & 1/3           & \textbf{1/3}         \\ \hline
\multirow{4}{*}{DeepOnet} & PSPO & \textbf{4.3882} & 0.1045          & \textbf{1.5018} & 0.0281          & 1.4397          & 0.0118          & \textbf{2/3}           & 0/3          \\& US            & 6.1094          & 0.1461          & 2.9384          & 0.1307          & 1.9674          & 0.0933          & 0/3           & 0/3          \\& CNS    & 4.9247          & \textbf{0.0660} & 1.5121          & 0.0414          & \textbf{1.0970} & \textbf{0.0098} & 1/3           & \textbf{2/3}          \\ & ECS & 5.6234          & 0.1041          & 1.5606          & \textbf{0.0218} & 1.1606          & 0.0101          & 0/3           & 1/3          \\ \hline
\multirow{4}{*}{PODNN}    & PSPO                      & 1.5541          & 0.0337          & 0.6914          & 0.0044          & 0.6451          & 0.0038          & 0/3           & 0/3          \\& US            & 2.7240          & 0.0526          & 2.5695          & 0.0187          & 1.0092          & 0.0038          & 0/3           & 0/3          \\& CNS    & \textbf{0.8210} & \textbf{0.0071} & \textbf{0.6116} & \textbf{0.0033} & \textbf{0.6081} & \textbf{0.0028} & \textbf{3/3}           & \textbf{3/3}          \\& ECS & 2.1037          & 0.0238          & 0.6123          & 0.0037          & 0.6401          & 0.0037          & 0/3           & 0/3          \\ \hline
\multirow{4}{*}{PINN}     & PSPO                      & 1.9292          & 0.1392          & 0.8718          & 0.0096          & 0.5343          & 0.0043          & 0/3           & 0/3          \\& US            & 4.6241          & 0.1609          & 2.2918          & 0.0157          & 0.8240          & 0.0053          & 0/3           & 0/3          \\& CNS    & \textbf{1.7246} & \textbf{0.1272} & 0.8937          & \textbf{0.0090} & 0.6258          & \textbf{0.0042} & 1/3           & \textbf{3/3}          \\& ECS & 2.3860          & 0.1439          & \textbf{0.8132} & 0.0099          & \textbf{0.4009} & 0.0043          & \textbf{2/3}           & 0/3          \\ \toprule
\end{tabular}
}
\label{tab:case2_num}
\end{table*}

\begin{figure*}[!h]
	\centering
\includegraphics[width=1.0 \linewidth]{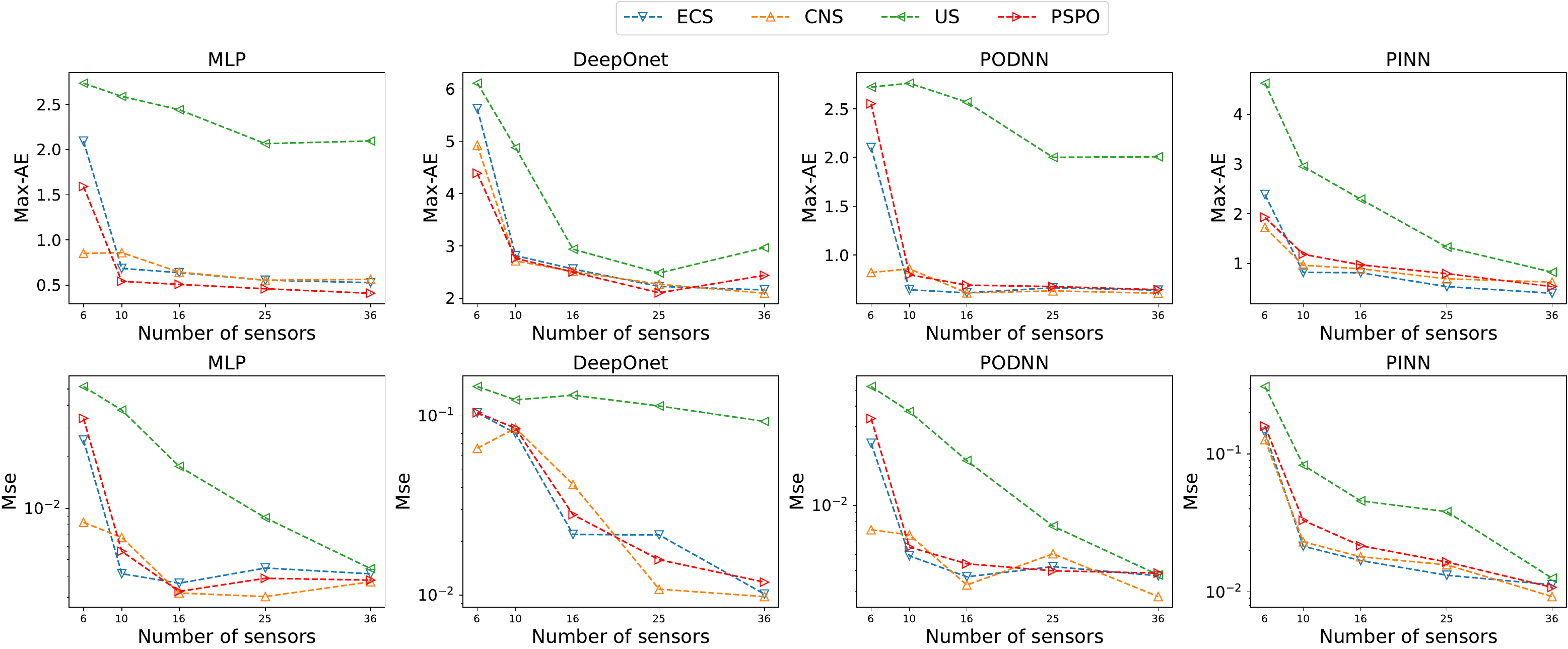}
	\caption{Application case: comparisons of MLP, DeepOnet, PODNN, and PINN with PSPO, US, CNS and ECS under three noise scales. The row represents predictions under $\lambda_1=0.49,\lambda_3=2.25$ and $\epsilon_{u} \sim \mathcal{N}\left(0, 0.01^{2}\right)$. The columns (from left to right) represent results of MLP, DeepOnet, PODNN, and PINN, respectively.}
	\label{fig:case2_num}
\end{figure*}

\vspace{7cm}
\subsubsection{Comparisons with data-driven sensor placement methods}
% On the premise that the large measured data cannot be obtained, such as the experimental design stage, data-driven sensor placement methods are not practical. However, 
To further investigate the effectiveness of PSPO, we assume there is sufficient data available to construct data-based criteria. PSPO is compared with CNS and ECS. Table \ref{tab:case2_num} shows comparisons by PSPO, US, CNS, and ECS under the different sensor numbers. MLP and DeepOnet by PSPO achieve the best Max-AE. Moreover, in most instances, models by CNS attain the best Max-AE and the best MSE, while models by PSPO can achieve close results. Furthermore, PSPO is superior to US under different sensor numbers. It is intuitive to get the more conclusion from Fig. \ref{fig:case2_num}. The trends of Max-AEs and MAEs by PSPO, CNS, and ECS are close. PSPO can reach or get close to the best Max-AE and the best MSE achieved by CNS or ECS. This indicates that models by PSPO gain a similar reconstruction performance as models by data-driven sensor placement optimization methods. Besides, Max-AEs and MAEs decrease with the increase of the sensor number, and the decrease becomes small for the sensor number $k>16$.

\begin{figure*}[!t]
	\centering
\includegraphics[width=1.0 \linewidth]{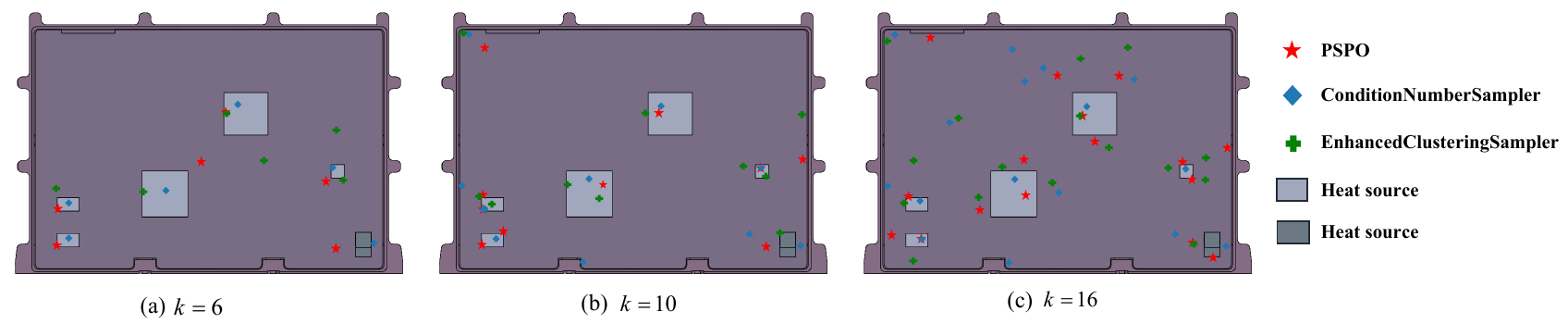}
	\caption{Application case: sensor locations generated by PSPO, CNS, and ECS under $k=6, 10, 16$.}
	\label{fig:case2_sensor_num_compare2}
\end{figure*}

\begin{figure*}[!h]
	\centering
\includegraphics[width=1.0 \linewidth]{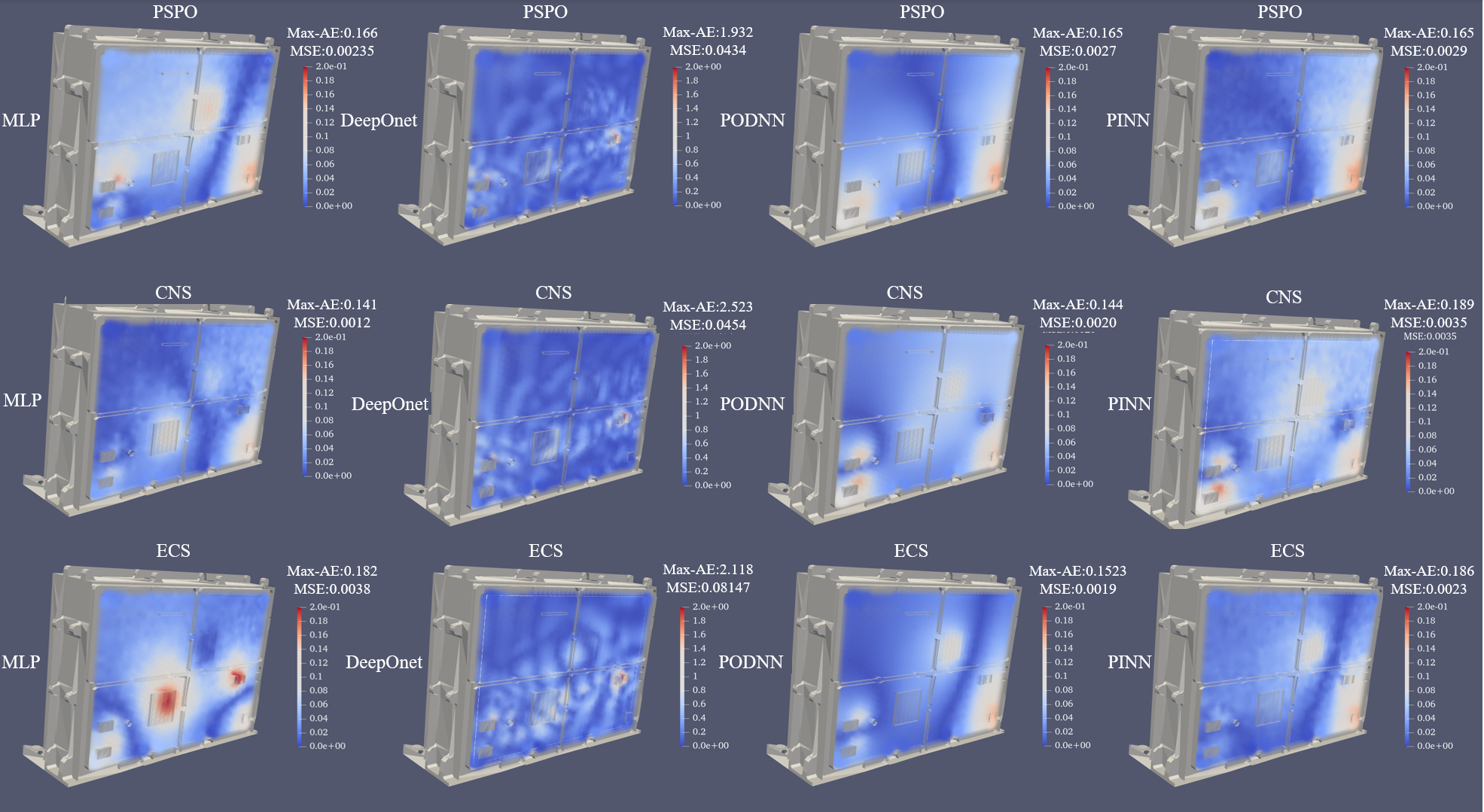}
	\caption{Application case: point-wise errors of MLP, DeepOnet, PODNN, and PINN using PSPO, CNS, and ECS under $k=10$ and $\epsilon_{u} \sim \mathcal{N}\left(0, 0.01^{2}\right)$. The columns (from left to right) represent results of MLP, DeepOnet, PODNN, and PINN, respectively. The rows represent results by PSPO, CNS, and ECS, respectively. It is worth noting that the maximum error interval is $2$ for DeepOnet and $0.2$ for other models.}
	\label{fig:case2_pred_compare2}
\end{figure*}

Fig.\ref{fig:case2_sensor_num_compare2} shows sensor locations generated by PSPO, CNS, and ECS under $k=6, 10, 16$. Under $k=6$, sensors by PSPO, CNS, and ECS are distributed around heat sources. For $k=10, 16$, most sensors by PSPO, CNS, and ECS are located near heat sources, with a few distributed near the boundary. Sensors near heat sources and the boundary provide more important physics information. Furthermore, sensors by PSPO, CNS, and ECS are close in some areas. This indicates that both the data-driven and physics-driven methods consider these areas to be critical for reconstruction. Fig. \ref{fig:case2_pred_compare2} illustrates point-wise errors by RS, CNS, and ECS, respectively. The point-wise errors are comparable by PSPO, CNS, and ECS. In addition, large error areas are mainly concentrated around heat sources. Therefore, more sensors placed around the heat source can provide more accurate heat source information to improve reconstruction performances.

% \vspace{4cm}
\subsubsection{Performance with different training data scales}
To further investigate reconstruction performances under different training data scales, we compare PSPO with CNS, ECS, RS, and US. CNS and ECS use training data of different scales to determine sensor locations. RS generates $20$ sets of sensor locations, and the best MSE is recorded. Besides, since the reconstruction performance of PINN is mainly affected by the sensor number and the noise scale, PINN is not considered in this part. The different training data scale is chosen from $\{200, 2000, 10000\}$. The number of test data is $50$. Training data and test data are generated by randomly sampling from problem parameters.

\begin{figure*}[!htbp]
	\centering
\includegraphics[width=0.93 \linewidth]{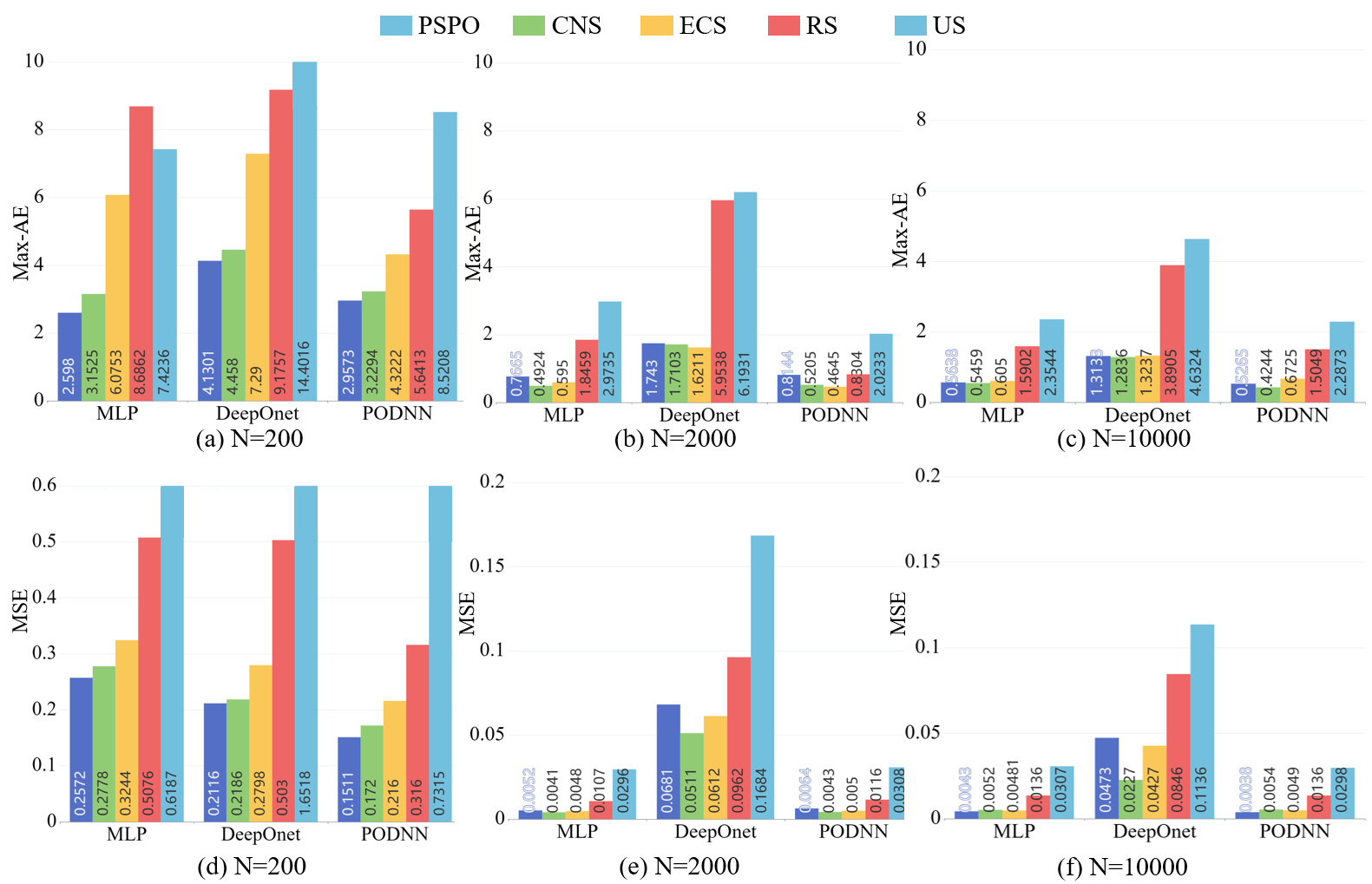}
	\caption{Application case: comparisons of MLP, DeepOnet, and PODNN using PSPO, CNS, ECS, RS, and US under different training data scales. The rows represents Max-AEs and MSEs under $\epsilon_{u} \sim \mathcal{N}\left(0, 0.01^{2}\right)$. The columns (from left to right) represent results of MLP, DeepOnet, and PODNN, respectively.}
	\label{fig:trainnum}
\end{figure*}

Fig. \ref{fig:trainnum} shows comparisons of MLP, DeepOnet, and PODNN using PSPO, CNS, ECS, RS, and US under different training data scales. As the figure shows, with more training data, Max-AE and MSE decrease in general. The reconstruction performances of models improve significantly when the training data increases from  $200$ to $2,000$, while Less improvement is observed when the training data scale from $2,000$ to $10,000$. In terms of Max-AE and MSE, PSPO, CNS, and ECS outperform RS and US under different training data scales. More importantly, models using PSPO are slightly superior to models using CNS and ECS under $N=200$. This indicates that CNS and ECS are limited by the number of training data. When the size of training data increases to 2000 or 10,000, CNS and ECS show significantly improved performance, slightly surpassing PSPO. This result is reasonable and convincing because PSPO optimizes placement based on physics-based criterion under data-free scenarios, while CNS and ECS use data-based criteria, making their advantages more pronounced with larger dataset. In particular, DSPO is always significantly superior to RS and US for DeepOnet, verifying the effectiveness of PSPO under data-free scenarios. PSPO performs sensor selection without a database (stricter constraints), so there is no problem even if DSPO is inferior to CNS for DeepOnet when a large number of training data is available.

\subsubsection{Performance with different parameters of neural network and GA.}
To further explore the impact of different neural network configurations, we deepen the network layers and widen the number of neurons in each layer based on the baseline. The results of this investigation are summarized in Table \ref{tab:net_case2}. Among all models, the baseline model consistently demonstrated superior performance, particularly in terms of Max-AEs and MSEs. When compared to the baseline, widening the network generally leads to an increase in both MSEs and Max-AEs. In contrast, deepening the network tends to increase MSEs but yields a reduction in Max-AEs. Notably, DeepONet exhibits worse performance under both network deepening and widening, with significant increases in both Max-AEs and MSEs. This indicates that there is a trade-off between deepening the network layers and widening the number of neurons to achieve superior reconstruction performance.

\begin{table}[!h]
\centering
\caption{Application case: comparisons of four models with different parameter setting. The sensor locations are obtained through DSPO. The noise level is $\varepsilon_u \sim \mathcal{N}\left(0,0.1^2\right)$. The number of observations is $6$. The best results under the different sensor number are highlight.}
\scalebox{0.86}{
\begin{tabular}{ccccc}
\toprule
\multicolumn{2}{c}{Model}            & Layers                                                                                                        & Max-AE         & MSE            \\ \hline
\multirow{3}{*}{MLP}      & Baseline & {[}6,64,64,64,1175{]}                                                                                         & 1.592          & \textbf{0.034} \\
                          & Deepen   & {[}6,64,64,64,64,1175{]}                                                                                      & \textbf{1.517} & 0.044          \\
                          & Widen    & {[}6,128,128,128,1175{]}                                                                                      & 1.896          & 0.042          \\ \hline
\multirow{3}{*}{PODNN}    & Baseline & {[}6,64,64,64,10{]}                                                                                           & 2.554          & \textbf{0.034} \\
                          & Deepen   & {[}6,64,64,64,64,10{]}                                                                                        & \textbf{1.556} & 0.044          \\
                          & Widen    & {[}6,128,128,128,10{]}                                                                                        & 1.631          & 0.042          \\ \hline
\multirow{3}{*}{PINN}     & Baseline & {[}2,64,64,64,1{]}                                                                                            & 1.930          & \textbf{0.136} \\
                          & Deepen   & {[}2,64,64,64,64,1{]}                                                                                         & \textbf{1.517} & 0.142          \\
                          & Widen    & {[}2,128,128,128,1{]}                                                                                         & 1.896          & 0.138          \\ \hline
\multirow{4}{*}{DeepOnet} & Baseline & \begin{tabular}[c]{@{}c@{}}Net1:{[}6, 128,128,128,128{]}\\ Net2:{[}2, 128,128,128,128{]}\end{tabular}         & \textbf{4.389} & \textbf{0.104} \\
                          & Deepen   & \begin{tabular}[c]{@{}c@{}}Net1:{[}6, 128,128,128,128,128{]}\\ Net2:{[}2, 128,128,128,128,128{]}\end{tabular} & 8.989          & 0.751          \\
                          & Widen    & \begin{tabular}[c]{@{}c@{}}Net1:{[}6, 256,256,256,256{]}\\ Net2:{[}2, 256,256,256,256{]}\end{tabular}         & 14.710         & 0.831          \\ \toprule
\end{tabular}
}
\label{tab:net_case2}
\end{table}

The comparisons of function and computational cost under different parameters of GA are shown in Table \ref{tab:GA_case2}. A larger population size provides greater diversity, potentially leading to better solutions. However, the increased population size also significantly raises computational cost, with the time required extending to 83.3 minutes. The higher crossover rate at $90\%$ maintains superior genes from parent individuals, resulting in better results. A higher mutation rate aids in exploring new solutions to avoid local optima, but if too high, it may disrupt the preservation of superior genes. Therefore, we have to set a moderate mutation rate. In conclusion, the results demonstrates the importance of balancing population size, crossover, and mutation rates to optimize both solution quality and computational efficiency.
    
\begin{table}[!h]
\centering
\caption{Application case: comparisons of function and computational cost under different parameters of GA. The sensor number is 16. The best function value are highlight.}
\scalebox{0.86}{
\begin{tabular}{cccccccccc}
\toprule
\multirow{2}{*}{Setting} & \multicolumn{3}{c}{Population size} & \multicolumn{3}{c}{Crossover rate} & \multicolumn{3}{c}{Mutation rate} \\ \cline{2-10} 
                         & 10       & 20              & 30     & 70\%    & 80\%   & 90\%           & 1/32    & 1/16            & 1/8    \\ \hline
$log(\kappa(\boldsymbol{W}))$                  & 9.15     & \textbf{9.08}   & 8.92   & 9.14    & 9.11   & \textbf{9.08}  & 9.12    & \textbf{9.08}   & 9.15   \\
Time/min                 & 28.51    & 57.95           & 83.3   & 55.58   & 57.11  & 56.87          & 57.46   & 58.87           & 57.15  \\ \toprule
\end{tabular}
}
\label{tab:GA_case2}
\end{table}

It's worth noting that although this study utilizes simulation data, we prefer to obtain global field data using infrared cameras. While the physics-based criterion is derived from RBF-FD, optimized placement by PSPO have transferability to improve reconstruction performance of NNs, which are trained with numerical and experimental data. The criterion through RBF-FD is grounded in PDEs, which share fundamental similarities with experimental and numerical data in describing thermal conduction phenomena. NN-based models with numerical or experimental data are also to learn the thermal conduction phenomena. This similarity forms the basis of their transferability for NNs. Our experiments have demonstrated that optimized locations by PSPO exhibit transferability to improve reconstruction performance of NNs with numerical data. Overall, our DSPO based on incomplete model for data-free scenarios provides valuable insights and guidance for practical engineering placement.

\section{Conclusion}
\label{sec conclusion}
This work proposes a novel physics-driven sensor placement optimization methodology to obtain the best sensor locations using a physics-based criterion in data-free scenarios, which is validated by the NN-based reconstruction models, including MLP, DeepOnet, PODNN, and PINN. The optimal solution of the reconstruction task is analyzed to derive the theoretical bounds of the reconstruction error under noise scenarios. It is proven that error bounds correlate with the condition number determined by sensor locations. The condition number as the physics-based criterion is to optimize sensor locations by the genetic algorithm. Finally, three types of NN-based models are used to validate the proposed method. Numerical cases, including a simple case and an application case, are utilized to demonstrate the efficiency of the PSPO method. The results show the PSPO method outperforms random and uniform selection methods, improving the reconstruction performances significantly. Besides, the PSPO method achieves reconstruction performances similar to existing data-driven placement optimization methods. On another note, while there are numerous data-driven methods for sensor selection, the data may be fraught with substantial noise or even corruption, casting doubt on the reliability of these methods for the selection of sensor positions. In contrast, we have proposed the PSPO approach to sensor selection, which introduces innovative theories and methodologies. Once the sensor locations are determined, it is relatively straightforward to select, replace, or update the reconstruction methods. However, if the selection of sensor locations is suboptimal due to inherent flaws in the data, revising the physical deployment of sensor hardware can be quite costly. Therefore, regardless of the reconstruction method used subsequently, our physics-driven approach to sensor selection is a reliable and prudent choice.

Although the PSPO method has exhibited some superior potential in data-free scenarios, we believe that perhaps a combination of data-driven and physics-driven methods could lead to a better approach for sensor placement selection, which is a research direction we intend to pursue in the future. Additionally, greedy algorithms are also efficient for single TFR tasks, providing quick and optimal solutions. We aim to further integrate physics-based criterion with greedy algorithms to explore placement selection in the future work.

\nolinenumbers

\section*{CRediT authorship contribution statement}
\textbf{Xu Liu}: Software, Methodology, Formal analysis, Writing - original draft. \textbf{Wen Yao}: Supervision, Funding acquisition, Project administration. \textbf{Wei Peng}: Supervision, Review \& Editing. \textbf{Zhuojia Fu}: Review. \textbf{Zixue Xiang}: Visualization. \textbf{Xiaoqian Chen}: Supervision, Project administration.

\section*{Declaration of competing interest}
The authors declare that they have no known competing financial interests or personal relationships that could have appeared to influence the work reported in this paper.

\section*{Data and Code availability}
The dataset and code generation scripts in this study are available at https://github.com/liuxu97531/PSPO.

\section*{Acknowledgements} 
This work was supported by National Natural Science Foundation of China under Grant No.52005505, and 62001502.

\bibliography{elsarticle-template}

\end{document}